\documentclass[sigconf]{acmart}
\usepackage{enumitem} 
\usepackage{amsmath}
\usepackage{multirow}
\usepackage{subfig}
\usepackage{graphicx}
\usepackage{subfloat}
\usepackage{pifont}
\AtBeginDocument{%
  }

\setcopyright{none}
\copyrightyear{2018}
\acmYear{2018}
\acmDOI{XXXXXXX.XXXXXXX}
\acmConference[Conference acronym 'XX]{Make sure to enter the correct
  conference title from your rights confirmation email}{June 03--05,
  2018}{Woodstock, NY}
\acmISBN{978-1-4503-XXXX-X/2018/06}

\begin{document}

\title{KairosVL: Orchestrating Time Series and Semantics for \\Unified Reasoning}

\author{Haotian Si}
\authornote{Also with the University of Chinese Academy of Sciences.}
\affiliation{%
  \institution{CNIC, CAS}
  \state{Beijing}
  \country{China}}

\author{Changhua Pei}
\authornote{Also with the Hangzhou Institute for Advanced Study, University of Chinese Academy of Sciences.}
\authornote{Corresponding Author. Email: chpei@cnic.cn}
\affiliation{%
  \institution{CNIC, CAS}
  \state{Beijing}
  \country{China}}

\author{Xiao He }
\affiliation{%
  \institution{ByteDance}
  \state{Beijing}
  \country{China}}

\author{Zeyan Li}
\affiliation{%
  \institution{ByteDance}
  \state{Beijing}
  \country{China}}

\author{Zhe Xie}
\affiliation{%
  \institution{Tsinghua University}
  \state{Beijing}
  \country{China}}

\author{Zexin Wang}
\affiliation{%
  \institution{CNIC, CAS}
  \state{Beijing}
  \country{China}}

\author{Jiyao Hu}
\affiliation{%
  \institution{ByteDance}
  \state{ San Jose}
  \country{USA}}

\author{Zhaoyang Yu}
\affiliation{%
  \institution{ByteDance}
  \state{Beijing}
  \country{China}}

\author{Tieying Zhang}
\affiliation{%
  \institution{ByteDance}
  \state{ San Jose}
  \country{USA}}

\author{Dan Pei}
\affiliation{%
  \institution{Tsinghua University}
  \state{Beijing}
  \country{China}}

\author{Jianhui Li}
\affiliation{%
  \institution{Nanjing University}
  \state{Nanjing}
  \country{China}}

\author{Gaogang Xie}
\affiliation{%
  \institution{CNIC, CAS}
  \state{Beijing}
  \country{China}}

\renewcommand{\shortauthors}{Trovato et al.}

\begin{abstract}
Driven by the increasingly complex and decision-oriented demands demands of time series analysis, we introduce the Semantic-Conditional Time Series Reasoning task, which extends conventional time series analysis beyond purely numerical modeling to incorporate contextual and semantic understanding. To further enhance the model’s reasoning capabilities on complex time series problems, we propose a two-round reinforcement learning framework: the first round strengthens the model’s perception of fundamental temporal primitives, while the second focuses on semantic-conditioned reasoning. The resulting model, KairosVL, achieves competitive performance across both synthetic and real-world tasks. Extensive experiments and ablation studies demonstrate that our framework not only boosts performance but also preserves intrinsic reasoning ability and significantly improves generalization to unseen scenarios. To summarize, our work highlights the potential of combining semantic reasoning with temporal modeling and provides a practical framework for real-world time series intelligence, which is in urgent demand. 

\end{abstract}



\maketitle


\section{Introduction}
Time series analysis is a foundational component of data science that underpinning a wide spectrum of applications ranging from AIOps for large-scale web service system reliability, to user behavior modeling, intrusions and fraud detection, and content recommendations~\cite{sec5,sec6,sec7,sec8,sce3,sce4,scene1,scene2}. For a long time, the academic and industrial focus on time series analysis has centered on a well-defined set of tasks, such as time series forecasting~\cite{fore1, fore2,fore3,fore4}, classification~\cite{cla1,cla2,cla3}, and anomaly detection~\cite{ano1,ano2,ano3,ano4}, in a numerical data processing paradigm. To this end, a vast literature of statistical and, more recently, deep learning methods has emerged. The advent of Large Language Models (LLMs) and Multimodal Large Language Models (MLLMs) has continued this trajectory, with current research exploring their application to these established tasks, either to improve pattern recognition performance by introducing more information, or to enhance the model's interpretability~\cite{tl1,tl2,tl3,tl4}. 

\begin{figure}[htb]
    \centering
    \includegraphics[width=\linewidth]{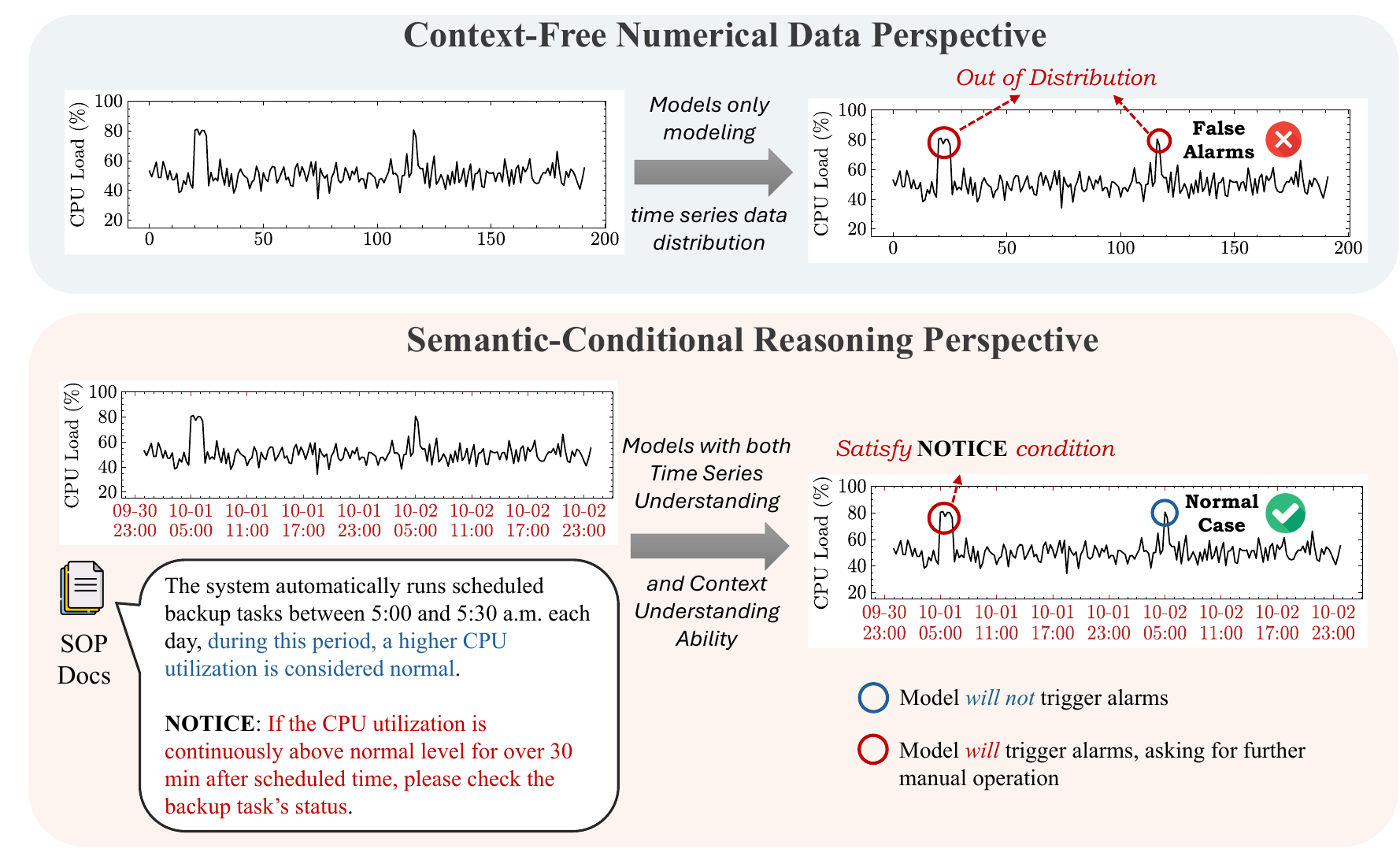}
    \caption{Comparison of time series task formulations from the perspectives of Numerical Data Processing and Semantic-Conditional Reasoning (taking the real-world time series anomaly diagnosis task as an example). While the majority of existing studies remain confined to the former, real-world applications also urge for the latter, an end-to-end reasoning paradigm.}
    \label{fig:motiv1}
\end{figure}

However, in complex operational environments, previous works are constrained by a prevailing pure numerical data processing perspective that is fundamentally insufficient for real-world challenges. In fact, effectively interpreting these data requires deep understanding of the application context, system architecture, and operational semantics. For example, in troubleshooting tasks in web services operations shown in Fig.~\ref{fig:motiv1}, a recurring daily spike in CPU utilization might be persistently flagged as an anomaly by a numerical value-based monitor, as this is an out-of-distribution (OOD) phenomenon. However, this pattern should be correctly identified as benign when contextualized with the operational knowledge that a data backup script is scheduled to run at the top of every day. 
Experiments in Appendix~\ref{app:tradits} show that without contextual grounding, such models misclassify benign events as anomalies, leading to incorrect or even misleading conclusions. 

Inspired by the above practical demand, we extend traditional numerical time-series analysis into a contextual reasoning paradigm, i.e., \textbf{Semantic-Conditional Time Series Reasoning}. Some existing research has shown that modern LLMs/MLLMs possess a degree of capability for simple time series comprehension, with vision-based MLLMs generally outperforming text-only LLMs~\cite{tl4,ben1-5, zhang2025timemaster,timerbed}. Yet, our empirical analysis reveals that these models fall short when confronted with the complex demands of semantic-conditional time series reasoning. Even state-of-the-art models consistently fail when evaluated on real-world tasks, for example, failing to focus on relevant time windows, ignoring or misinterpreting scheduled events, and hallucinating pattern that never occurs. Furthermore, we observe that some failure cases even stem from lacking accurate perception of basic temporal properties—misestimating the period of traffic flow as 20 hours instead of 24 hours and resulting in wrong prediction. This demonstrates that without a robust understanding on intrinsic temporal characteristics, higher-order reasoning tasks inevitably collapse. Taken together, these findings reveal that existing models still struggle with real-world time series reasoning tasks, highlighting an urgent need for novel training frameworks that can systematically enhance models’ temporal reasoning capabilities.

To address the above challenge, in this paper, we propose a two-round reinforcement learning with verifiable reward (RLVR) framework, where each round is paired with carefully constructed tasks and alignment settings, to systematically facilitate the model's ability on real-world time series reasoning. In the initial round, we post-train MLLM on rule-driven datasets composed of basic time series primitives. This enhances the model's grasp of core time series concepts such as trends, seasonality, outliers, and noise levels, and further strengthens its capability to align visual features with their corresponding timestamps, which establishs a perceptual foundation for subsequent reasoning tasks. In the second round, we deconstruct the complex semantic-conditional time series reasoning tasks into four core subtasks observed in real-world applications: factual-adherent task, predictive task, event-aware task, and counterfacutal task. We then construct a diverse, high-quality reasoning dataset containing all subtask using an LLM-enhanced data construction pipeline, KairosDataPipe, followed by rigorous human verification. The model undergoes its second round of RL training on this specialized dataset to master these advanced reasoning skills. Experimental results shows that \textbf{KairosVL}, the model post-trained under our proposed frameworks based on Qwen2.5VL-7B~\cite{qwen2.5vl}, achieves competitive performance compared with leading commercial models not only on \textbf{KairosBench} which is constructed based on real-world demands, but also on zero-shot real-world AIOps fault diagnosis samples. 
Ablation studies confirm that the proposed two‑round curriculum substantially improves perceptual accuracy, reasoning robustness, and generalization.

\noindent This paper makes the following contributions:
\begin{enumerate}[leftmargin=*,align=left]
    \item We pioneer systematically defining the Semantic-Conditional Time Series Reasoning task 
    , which extends the boundary of existing time series analysis tasks.
    \item We propose a two-round reinforcement learning framework that progressively enhances the model’s perception and reasoning abilities, and our model KairosVL achieves competitive performance across both synthetic and real-world time series reasoning tasks compared with baselines.
    \item Ablation studies demonstrate that our framework not only boosts the model’s performance, but also effectively preserves its intrinsic reasoning ability and significantly improves generalization on real-world samples.
\end{enumerate}

\section{Preliminaries and Related Works}
\label{sec:def}
In this section, we introduce the definition of Semantic-Conditional Time Series Reasoning task and some potential paths towards building time series reasoning model. Some related works about existing time series reasoning tasks are discussed in Appendix~\ref{app:task}.
\subsection{Problem Definition}
Previous \textbf{numerical time series data analysis} treats time series exclusively as a sequence of numerical measurements. Its objective is to learn a mathematical function function $\mathcal{F}: \mathbb{X} \rightarrow \mathbb{Y}$ that executes a direct mapping from a purely numerical input space $\mathbb{X} \subseteq \mathbb{R}^T$ to a target space $\mathbb{Y}$. Notably, $\mathbb{Y}$ is a space of atomic, numerical-level answers. An element $y\in\mathbb{Y}$ represents a direct conclusion (e.g., a class label $\{0,1\}$ or a vector in $\mathbb{R}^k$. In contrast, \textbf{semantic-conditional time series reasoning tasks} emphasize modeling temporal data at the level of natural language semantics with LLMs/MLLMs, while highlighting the role of contextual information. Formally, let $X\in \mathbb{X}$ be the time series paid attention to and $\mathcal{H}(\cdot)$ be an operation to convert time series data to the format that can be understood by LLMs/MLLMs (text/image/TS embedding). Let $K\in \mathcal{K}$ be pieces of contextual knowledge from a natural language knowledge space $\mathcal{K}$(which may contain rules, text, or structured data). The objective is to learn a language generation model, $\mathcal{G}$, that reasons over both the time series data $X$ and the context $K$:

$$\mathcal{G}: \mathcal{H}(\mathbb{X}) \times \mathcal{K} \rightarrow \mathbb{Y'}|\mathbb{Z}$$

Here, different from $\mathbb{Y}$, $\mathbb{Y'}$ denotes the space of semantic-level matching, e.g., natural language choices, or statements whose correctness is judged by text extraction. $\mathbb{Z}$ denotes the interpretation space. An element $z\in\mathbb{Z}$ indicates one reasoning trajectory that explains why the certain conclusion $y'$ is reached. We introduce $\mathbb{Z}$ in the formulation because it makes the entire decision-making process generated by $\mathcal{G}$ traceable from input to output and allowing for systematic debugging, which is crucial for real-world applications.

\subsection{Time Series Reasoning Models}
The continuous emergence of Large Language Models and Multimodal Models has demonstrated increasingly powerful semantic reasoning capabilities. Existing research attempts to align the time series modality with the text modality of these models through various paradigms to better accomplish time series reasoning tasks. These paradigms can be broadly categorized into the following three types:
\begin{itemize}[leftmargin=*,align=left]
\item LLMs. This paradigm~\cite{tl4} converts raw time series into a list of numerical text, which is then fed into the model. This allows the LLM to directly leverage its text comprehension abilities to understand the time series.
\item Vision-based MLLMs. This paradigm~\cite{timerbed,zhang2025timemaster} transforms raw time series into visualized line charts and input them into MLLMs, which rely on visual encoders to capture temporal patterns and align them with the textual modality.
\item TSLMs. These methods~\cite{xie2024chatts,wang2025itformer} design dedicated time-series encoders and achieve native alignment with the textual modality via time series–text pretraining.
\end{itemize}

Several works has experimentally demonstrated that Vision-based MLLMs tends to exhibit better performance and efficiency on existing time series analysis tasks against LLMs and TSLMs~\cite{ben1-5, zhang2025timemaster,timerbed,MCQ2,zhou2024can}, since LLMs often suffer from the excessive token cost and severe hallucinations caused by the overlong time series text, and TSLMs are often constrained by the scarcity and narrow coverage of series-text pairs that used for pretraining. \textit{Building on the experimental findings of existing works and the demands of practical applications, our work will focus on further enhancing the reasoning capabilities of MLLMs for semantic-conditional time series analysis tasks}. Further details are provided in Appendix~\ref{app:text}.

\begin{figure*}[tb]
    \centering
    \includegraphics[width=\linewidth]{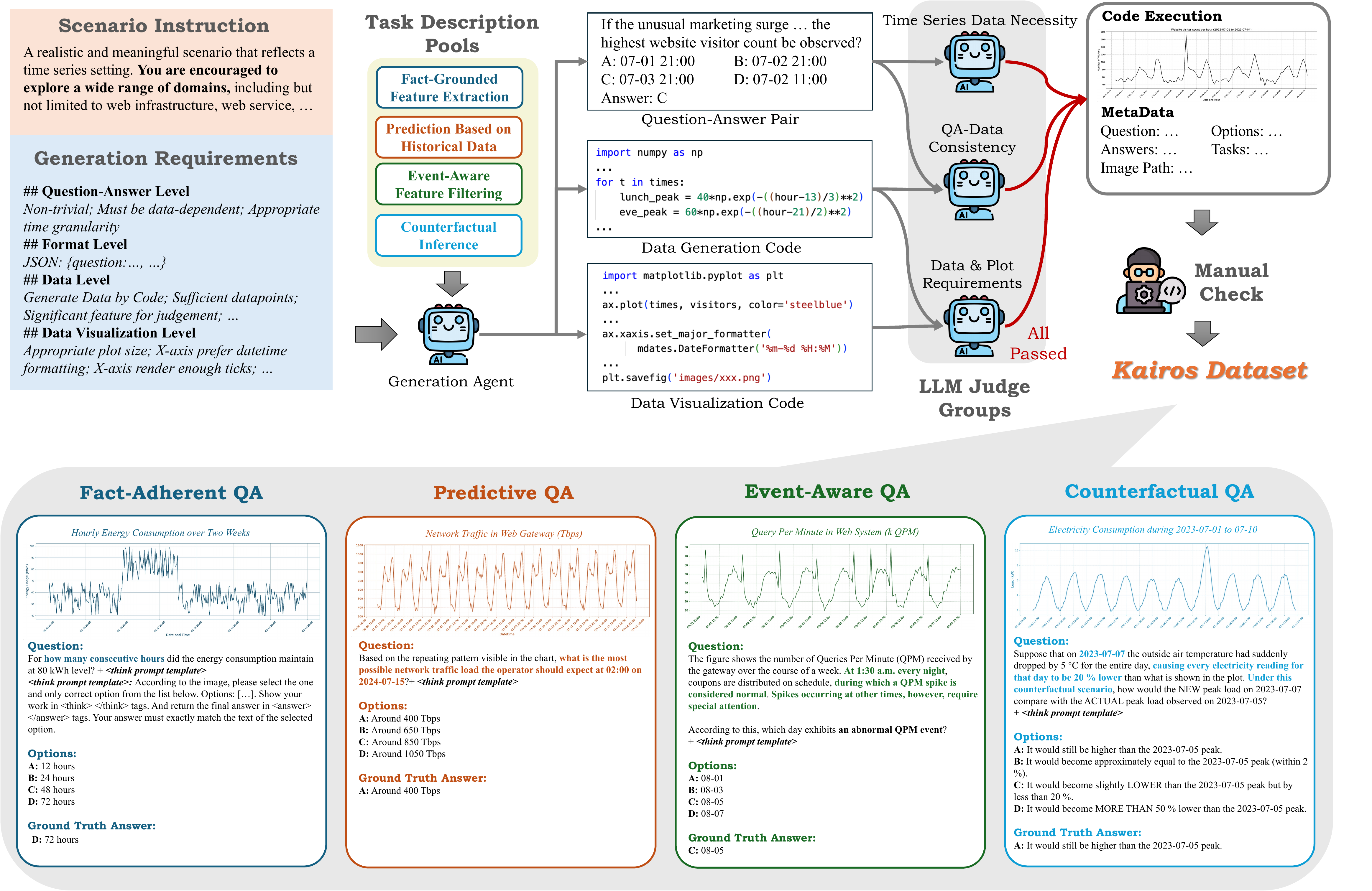}
    \caption{Details of KairosDataPipe and KairosDataset.}
    \label{fig:dataset}
\end{figure*}

\section{KairosDataPipe \& KairosVL}

Despite the urgent demand in industry for models capable of Semantic-Conditional Time Series Reasoning, which better aligns with the complex requirements of real-world time series analysis, existing academic research exhibits significant systematic gaps in this domain. To date, there has been no systematic formulation of this problem, let alone the development of corresponding benchmark datasets to support rigorous empirical evaluation. To bridge the gap between existing works and practical needs, we first introduce a structured perspective by categorizing Semantic-Conditional Time Series Reasoning into four subtasks, each emphasizing distinct aspects of reasoning demand. Building on this taxonomy, we propose KairosDataset, a high-quality time-series reasoning dataset generated through a carefully designed LLM-engaged multi-agent system, which provides large volumes of reasoning-focused samples to support post-training on existing MLLMs to further enhance their temporal reasoning capabilities and meet the demands of real-world applications.

\subsection{Taxonomy}
To provide a systematic formulation of the problem, we categorize Semantic-Conditional Time Series Reasoning into four subtasks:

\subsubsection{Fact-Grounded Feature Extraction.} This category focuses on extracting features and reasoning over factual properties explicitly observable from the time series data. Models are required to faithfully follow the underlying data patterns in the line chart rather than relying on priors or heuristics. Typical cases include but not limited to:
\begin{itemize}[leftmargin=*,align=left]
\item Identifying when a metric crosses a predefined threshold (e.g., "At which time does the value first exceed 100?").
\item Measuring the duration of specific conditions (e.g., "How long did the system remain above the safety limit?").
\item Comparing magnitudes across intervals (e.g., "Which quarter had the largest increase relative to the previous one?").
\end{itemize}

\subsubsection{Predictive Tasks.} The predictive dimension emphasizes forward-looking reasoning, requiring models to infer whether future points in the time series will meet certain conditions, often based on historical trends, seasonality, or cyclicity. Example tasks include but not limited to:
\begin{itemize}[leftmargin=*,align=left]
\item Estimating whether a variable will surpass a threshold within the next time window.
\item Anticipating periodic patterns such as weekly peaks.
\end{itemize}

\subsubsection{Event-Aware Feature Filtering.} In real-world scenarios, time series rarely exist in isolation; they must be interpreted in the context of external events and semantic conditions. This subtask evaluates a model’s ability to incorporate such contextual information and filter out irrelevant or misleading fluctuations. Typical cases include but not limited to:
\begin{itemize}[leftmargin=*,align=left]
    \item Excluding anomalies caused by external disruptions (e.g., "Ignore the temporary spike due to a scheduled system update and assess the underlying trend").
    \item Making conditional judgments (e.g., "If weekends are excluded, does the metric still exhibit a downward trajectory?").
\end{itemize}

\subsubsection{Counterfactual Inference.} The counterfactual dimension goes beyond factual and predictive reasoning, requiring models to simulate what-if scenarios by reasoning about alternative conditions not directly observed in the data. This is crucial for tasks such as causal analysis, stress testing, and decision support. Representative cases include:
\begin{itemize}[leftmargin=*,align=left]
    \item Hypothetical interventions (e.g., “If the cluster is expanded at time T, causing resource usage to drop by 20\% thereafter, when is the peak resource utilization observed within the year?”).
    \item Comparative scenarios (e.g., “If the growth rate after Q2 had remained the same as in Q1, what would the value be at year-end?”)
\end{itemize}

\subsection{KairosDataPipe}
An obstacle for enhancing the MLLM's capability on semantic-conditional time series reasoning is the lack of a large-scale high-quality structured reasoning dataset. To this end, we build \textbf{KairosDataPipe}, a LLM-engaged multimodal time series reasoning data generation pipeline as shown in Fig.~\ref{fig:dataset}. Due to the severe coherence issues of existing models in generating long-sequence time series data as text, as well as the significant hallucinations that arise when directly generating charts (such as unreadable or inconsistent labels), following and extending TSandLanguage~\cite{MCQ2}, KairosDataPipe adopts a "Text ($\mathcal{K}$ mentioned in Sec.~\ref{sec:def})--Data Generation Code (generating $\mathbb{X}$)--Visualization Code ($\mathcal{H}(\cdot)$)" sample generation schema. As shown in Fig.~\ref{fig:dataset}, we prompt GPT-o3 to generate the question-answer pairs, the corresponding python scripts of data generation, and the python scripts of data visualization code in diverse formats. We further design a multi-stage generation data check pipeline. All above improve not only the diversity and complexity of the generated samples, but also the data quality which helps to decrease the manual verification workload required to filter out the same quantity of valid samples. After manual check on each generated sample, we obtain \textbf{KairosDataset}, including all time series reasoning tasks mentioned above. Here are the key components of KairosDataPipe, and all prompts used in this pipeline are listed in Appendix~\ref{app:promptpipe}:

\subsubsection{Data Generation Agent.} 
We generate task samples by combining instruction templates with task descriptions drawn from reasoning task pools. Within the instruction templates, to enhance the diversity of scenarios represented in the dataset, we encourage the LLM to freely predefine a scenario. Based on this predefined scenario and a randomly selected task description, the LLM is then prompted to produce a time-series reasoning question–answer pair, the corresponding data generation code that matches the QA description, and the visualization code for rendering the generated data as a line chart. To maximize the usability of the generated samples, as illustrated in Fig.~\ref{fig:dataset}, we embed in the prompt a set of detailed requirements spanning multiple aspects, covering the quality assurance of QA pairs, the formatting of data, the specifications for data generation code, and the conventions for visualization scripts. 

\begin{figure*}[tb]
    \centering
    \includegraphics[width=0.95\linewidth]{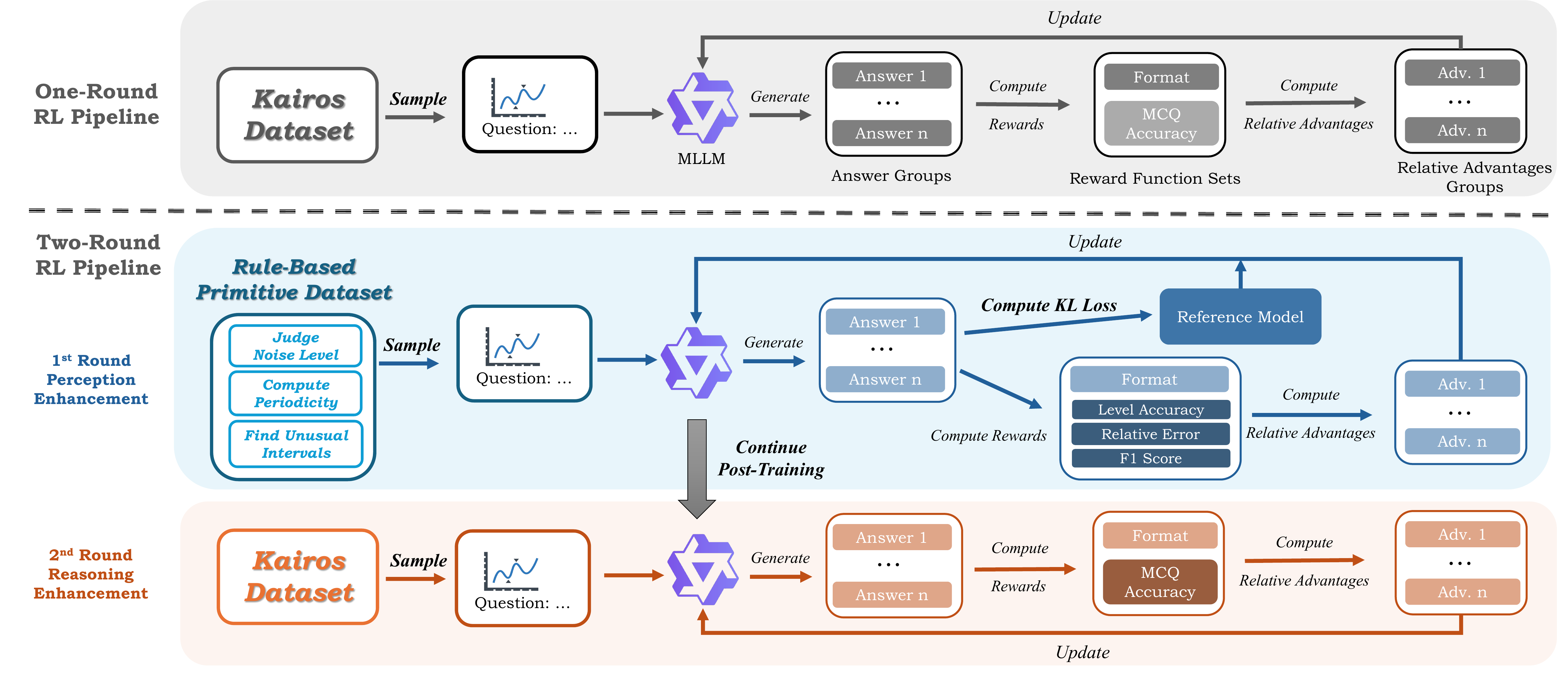}
    \caption{One-round reinforcement learning pipeline v.s. two-round reinforcement learning pipeline.}
    \label{fig:vl}
\end{figure*}

\subsubsection{LLM Judge Groups.}
During the initial stage of data generation, we observed that existing LLMs still have limited instruction-following abilities, often failing to fully satisfy the requirements specified in the prompt and thus producing unreasonable samples, such as the sames situation mentioned in ~\cite{MCQ2} where answers can be inferred directly from the question text without relying on the generated data. To mitigate this issue, we introduced a LLM-engaged judge group~\cite{llm-as-judge} including: the \textbf{Data Necessity Judge}, which attempts to answer the question solely from the textual description and flags the sample as "lacking sufficient data necessity" if the answer accuracy over five trials exceeds 50\%; the \textbf{QA–Data Consistency Judge}, which verifies whether the generated question and the associated data generation code lead to a correct and unique answer, ensuring consistency between the QA pair and its underlying data; and the \textbf{Data \& Plot Requirement Judge}, which re-examines the data generation and visualization code against the requirements, filtering out samples that violate structural or presentation constraints (e.g., overly sparse data points, improper time formats, or x-axes that are unintelligible for human readers). A sample is retained only if it passes all three judges. In practice, this multi-judge framework increases the proportion of valid samples from around 50\% to over 80\%, substantially improves the efficiency of subsequent human checking process.

\subsubsection{Plot Generation \& Manual Check.} 
For the samples that pass the judge group screening, we execute their data generation code and data visualization code to produce the corresponding text–image pairs. The corresponding code-to-plot python function is listed in Appendix~\ref{app:code1}. Each pair then undergoes meticulous human verification to guarantee both correctness and readability, ultimately yielding \textbf{KairosDataset}, a high-quality dataset containing over 2k time-series reasoning samples.

\subsection{KairosVL}

Several studies have demonstrated the effectiveness of Reinforcement Learning with Verifiable Reward (RLVR) ~\cite{guo2025deepseek} in the post-training phase of existing LLMs/MLLMs, showing that it enables models to generate more coherent reasoning trajectories, improves the accuracy of final answers, and enhances generalization across diverse scenarios, which tends to be a better choice compared with Supervised Fine-Tune (further discussion is listed in Appendix~\ref{app:sft}). Following these works, in our initial attempt we directly utilized KairosDataset and adopted Group Relative Policy Optimization (GRPO) ~\cite{grpo} with token-level loss optimization~\cite{yu2025dapo, drgrpo} and to conduct one round of reinforcement learning–based post-training, as shown in Fig.~\ref{fig:vl}. However, the model trained in this manner exhibited several issues (details listed in Sec.~\ref{sec:exphase} and Sec.~\ref{sec:excase}): 

\begin{enumerate}[leftmargin=*,align=left]
\item The model often misinterpreted certain fundamental temporal features during thinking, for example, mistaking a period of sustained high noise for a spike—leading to erroneous judgments. 
\item The reasoning length was overly short, with the generated reasoning offering mostly broad and generic descriptions rather than closely engaging with the underlying time-series information, which cannot provide helpful analysis reference.
\end{enumerate}

These issues reveal a fundamental challenge that the model’s poor grasp of fundamental time-series features greatly hinders its final reasoning performance, especially under reinforcement learning framework where model has to solely explore the correct answer according to their knowledge. Specifically, in the early stages of training, the model struggles to correctly perceive some basic temporal features like noise level, periodicity, or some significant out-of-distribution segments. As a result, reasoning chains that attempt to engage deeply with the time series are often less accurate than vague, generic descriptions. Consequently, the policy optimization process favors the latter, reinforcing shallow reasoning patterns instead of encouraging precise, temporally grounded analysis. This ultimately constrains its upper performance bound.

To overcome this issue, we construct an additional dataset aimed at aligning the model with basic temporal primitives, and \textbf{build a two-round reinforcement learning pipeline}. In the first round, the model undergoes reinforcement learning on this dataset, which helps it internalize and refine its perception of essential temporal structures. With this stronger foundation in place, we then introduce the KairosDataset for a second round of reinforcement learning, allowing the model to further strengthen its temporal reasoning abilities and tackle more challenging tasks with greater accuracy.

\subsubsection{Primitive Dataset.}
We curated an auxiliary dataset focusing on several critical basic time-series characteristics. The samples are generated by rules to ensure their correctness. Specifically, this dataset encompasses tasks including:
\begin{itemize}[leftmargin=*,align=left]
    \item Judging noise level (assessing the intensity of random signal variance);
    \item Judging if there is periodicity, and further identifying repeating patterns and quantifying their period length;
    \item Identifying unusual or significant out-of-distribution (OOD) patterns, providing their exact occurrence time and duration when they exist.
\end{itemize}

\subsubsection{Reward Function.}
Due to the introduction of different kinds of tasks, the reward function consists of a format part and several task-specific parts. If we cannot extract text within <answer></answer> tag, the format reward $r^{fmt}$ is set -0.5, otherwise $r^{fmt}$ is set to 0. This guides the model to produce outputs that strictly adhere to a predefined structure, which helps to easily extract the exact answer. For noise judgment task, we employ a classification reward that compare the selected item in [high, medium, low] with the ground truth. For period-related task, we employ a two phase reward: first judge if periodicity exists, and if exists, then compute the relative error between the model's answer and the ground truth as additional reward. For OOD task, we first judge if the OOD phenomenon exists, and if exists, we calculate the F1 score based on the given OOD intervals and the ground truth intervals as reward. For Multi-choice question tasks in KairosDataset, we directly compute the choice accuracy as reward. Formally, suppose the model's answer $\hat{c}$ is extracted within <answer></answer> tag, and given the ground truth answer $c$, the task specific reward $r^{task}$ and the overall reward $r$ can be calculated by:
\begin{equation}
    r^{task} =
\begin{cases}
\mathbb{I}[c=\hat{c}], & \text{if task in [\textit{Noise}, \textit{MCQ}]} \\
\min(\mathbb{I}[c_{e}=\hat{c_{e}}], 1-\frac{c_{p} - \hat{c_{p}}}{\hat{c_{p}}}), & \text{if task is \textit{Periodicity}} \\
\min(\mathbb{I}[c_{e}=\hat{c_{e}}], F1(c_{i}, \hat{c_{i}})), & \text{if task is \textit{OOD Detection}}
\end{cases}
\end{equation}

\begin{equation}
    r = \min\{r^{fmt}, r^{task}\}
\end{equation}
where $\mathbb{I}[\cdot]$ denotes the indicator function, $c_{e}$ denotes the feature existence part, $c_{p}$ denotes the period value part, and $c_{i}$ denotes the OOD intervals part.

\subsubsection{$1^{st}$ Round Perception Enhancement}
The first round reinforcement learning is dedicated to bootstrapping the model's foundational understanding of temporal primitives using Primitive dataset. Since the reasoning steps for these tasks are typically more constrained and less open-ended, there exists a risk that the model may overfit to these narrow objectives, potentially degrading its reasoning ability on more complex time-series tasks. Thus in this round, we incorporate a KL divergence regularization term with respect to a reference model (Qwen2.5VL-7B) into the training loss to prevent the policy from collapsing into overspecialized solutions. The training loss in the first round is formulated as follows:
\begin{multline}
J_{1^{st}}(\theta) = \mathbb{E}_{\{ts,q\}\sim P(TS,Q),\{o_i\}_{i=1}^G\sim\pi_{\theta_{old}} (O|ts,q)} \\
\left\{
\begin{aligned}
\frac{1}{\sum_{i=1}^G|o_i|}\sum_{i=1}^G\sum_{t=1}^{|o_i|}
\left\{
\begin{aligned}
\min [clip\left(r_{i,t}(\theta),1-\epsilon,1+\epsilon\right)A_{i,t}, \\r_{i,t}(\theta)A_{i,t}] -\beta\mathbb{D}_{KL}\left[\pi_\theta|| \pi_{ref}\right]
\end{aligned}
\right\}
\end{aligned}
\right\}
\end{multline}

where $P(TS,Q)$ denotes the time series plot-question pairs, $O$ denotes the text generated by the model, $\pi$ denotes the policy (i.e., the model), $r_{i,t}(\theta)=\frac{\pi_{\theta}(o_{i,t}|ts,q,o_{i,<t})}{\pi_{old}(o_{i,t}|ts,q,o_{i,<t})}$ is the importance sampling ratio for each generated token, $A_{i,t}=\frac{r_i-\text{mean}(\textbf{r})}{\text{std(\textbf{r})}}$, $\textbf{r}=\{r_1, r_2, \cdots,r_G\}$ denotes all rewards in the same group, $\epsilon$ is the clipping threshold, $G$ is the count of groups in the batch, and $\beta$ controls the strength of KL regularization.

\subsubsection{$2^{nd}$ Reasoning Enhancement.}
In the second round, we conduct continuous post-training on the checkpoint obtained by the first round training, to further enhance the model’s temporal reasoning capabilities on more complex tasks. Inspired by ~\cite{yu2025dapo}, we disable the KL divergence constraint, meanwhile increasing the upper clip value, to encourage the model to explore a wider range of reasoning trajectories, thus developing longer, more precise reasoning chains, grounded in temporal patterns, rather than producing shallow or overly conservative outputs. The training loss in the second round is formulated as follows:
\begin{multline}
J_{2^{nd}}(\theta) = \mathbb{E}_{\{ts,q\}\sim P(TS,Q),\{o_i\}_{i=1}^G\sim\pi_{\theta_{old}} (O|ts,q)} \\
\left\{
\begin{aligned}
\frac{1}{\sum_{i=1}^G|o_i|}\sum_{i=1}^G\sum_{t=1}^{|o_i|}
\min \left[clip\left(r_{i,t}(\theta),1-\epsilon,1+\epsilon'\right)A_{i,t}, r_{i,t}(\theta)A_{i,t}\right]
\end{aligned}
\right\}
\end{multline}

where $\epsilon'$ denotes the upper clip value that satisfies $\epsilon' > \epsilon$.

\subsubsection{Training Details.} 
Due to space limitations, please refer to the Appendix~\ref{app:train} for training settings and reward traces.

\section{Evaluation}

\begin{table*}[tb]
\renewcommand{\arraystretch}{1.05}
\tabcolsep=0.2cm
\centering
\caption{Overall Question-Answer Accuracy (\%) of different models on both Real-World datasets (Dataset $\mathcal{A}$) and KairosBench (Dataset $\mathcal{B}$). The highest, the second-best scores are shown in bold and underlined, respectively. For each sample in the dataset, we sample the answer five times and compute the mean accuracy value. The margin of errors can be found in Appendix~\ref{app:margin}.} 
\label{tab:mainres}
\resizebox{\textwidth}{!}{
\begin{tabular}{c|c|ccc|cccc}
\toprule[1.5pt]
\multirow{2}{*}{Method}&\multirow{2}{*}{Open-Source}& \multicolumn{3}{c|}{Dataset $\mathcal{A}$}&\multicolumn{4}{c}{Dataset $\mathcal{B}$}                             \\ \cmidrule{3-9}
      & &Scenario \#1&Scenario \#2&Scenario \#3&Fact-Adherent&Predictive&Event-Aware&Counterfactual \\ \midrule[1.2pt]
InternVL3-8B&\ding{51}&57.4&53.0&62.0&41.3&53.6&40.4&42.7           \\
Qwen2.5VL-7B&\ding{51}&75.6&53.3&45.3&47.6&50.0&36.9&42.6           \\
Qwen2.5VL-32B&\ding{51}&79.6&56.8&46.7&61.9&51.9&49.6&46.2           \\
Qwen2.5VL-72B&\ding{51}&{\underline {87.8}}&60.0&70.7&{\underline {71.2}}&42.4&54.4&\textbf{64.7}  \\
GLM4V-Plus&\ding{55} &66.7&{\underline {70.0}}&49.3&60.8&54.0&50.0&50.5           \\
GPT-4o&\ding{55} &83.3&64.5&\textbf{77.3}&57.1&63.3&58.3&{\underline {56.9}}     \\
Gemini-2.5Flash&\ding{55} &79.4&62.5&48.7&65.6&55.0&52.9&53.1           \\ \midrule[1.2pt]
KairosVL-\textit{SFT} (7B)& - &81.8&60.9&58.2&66.8&73.6&64.9&51.6          \\
KairosVL (7B)&\ding{51}&\textbf{88.9}&\textbf{70.4}&{\underline {73.3}}&\textbf{72.5}&\textbf{76.2}&\textbf{71.8}&56.4 \\
+Imp. (v.s. Qwen2.5VL-7B)&-&17.6\%&32.1\%&61.8\%&52.3\%&52.4\%&94.6\%&32.4\%         \\ \bottomrule[1.5pt]
\end{tabular}
}
\end{table*}

In this section, we conduct a comprehensive evaluation of KairosVL by addressing the following research questions (RQs):
\begin{itemize}[leftmargin=*,align=left]
    \item \textbf{RQ 1:} Can KairosVL achieve good performance on Semantic-Conditional time-series reasoning tasks, and does this capability generalize to real-world scenarios?
    \item \textbf{RQ 2:} Does the proposed Two-Round Reinforcement Learning Pipeline effectively enhance the model’s upper bound performance on time-series reasoning tasks?
    \item \textbf{RQ 3:} Can KairosVL provide clear, faithful, and logically consistent reasoning steps that not only improve answer correctness but also enhance interpretability and trustworthiness in practical applications?
\end{itemize}

\subsection{Experimental Setup}
In this section we will introduce the dataset for evaluation, the evaluation metric, and baselines. The prompt template used for training and test is listed in Appendix~\ref{app:prompt}.

\subsubsection{Evaluation Dataset.}
To comprehensively evaluate the performance of KairosVL and other methods on semantic-conditional time series reasoning tasks, we conduct evaluations on two datasets. 

Dataset $\mathcal{A}$ consists of around 100 real-world samples collected from production environments with human labels, covering three representative scenarios. Each scenario includes multiple fault escalation rules (described in SOP (Standard Operation Procedure) documents in the format of nature language) and the corresponding time series metrics snapshots collected at different time points. The operator needs to decide whether to escalate the alert priority, maintain the current priority, or mitigate the alert based on the current time series data and the description of the escalation/mitigation rules. Since the construction process of KairosDataset (including task, prompt, data) is entirely independent from this dataset, the evaluation on this dataset can effectively reflect the generalization (Zero-shot) capability of KairosVL.

Dataset $\mathcal{B}$, \textbf{KairosBench}, is a large-scale benchmark containing about 600 samples constructed with KairosDataPipe, but specifically designed for evaluation using more targeted prompts that reflect real-world requirements. In particular, we extract and summarize the temporal feature reasoning capabilities expected in real-world SOP samples and incorporate these insights into the prompts. This ensures that KairosBench keeps distinct from KairosDataset, while faithfully adhering to our task taxonomy.

\subsubsection{Evaluation Metric.}
The format of both Dataset $\mathcal{A}$ and Dataset $\mathcal{B}$ is multiple-choice. Each sample in $\mathcal{A}$ contains two options, whereas most samples in $\mathcal{B}$ contain four options, with a small number of samples containing two options. Therefore, we use the accuracy of multiple-choice question (MCQ) answers as the evaluation metric.

\subsubsection{Baselines.}
We compare our KairosVL against a wide range of open-source MLLMs across various size, including Qwen2.5VL-7B~\cite{qwen2.5vl}, InternVL3-8B~\cite{internVL3}, Qwen2.5VL-32B~\cite{qwen2.5vl}, and Qwen2.5VL-72B~\cite{qwen2.5vl}. We also compare KairosVL against some commercial MLLMs like GPT-4o~\cite{gpt4-o}, Gemini-2.5 Flash~\cite{gemini}, and GLM4V-Plus~\cite{glm2024chatglm}.
We exclude the comparison with LLMs (described in Section 2.3) since our initial experiments indicated their both inferior performance and substantial token consumption on time series reasoning tasks. Related information is listed in Appendix~\ref{app:text}. In addition to the models mentioned above, we also compare a variant that trained using SFT in both stages, KairosVL-\textit{SFT}, to intuitively illustrate the impact of post-training strategies on model performance.

\subsection{Overall Performance}
We list the evaluation results of KairosVL and other baselines in Table~\ref{tab:mainres}. On Dataset $\mathcal{A}$ containing real-world samples, KairosVL with enhanced reasoning capabilities demonstrates significant accuracy improvements across all scenarios compared to the base model Qwen2.5VL. Notably, in Scenario \#3, where most models perform poorly, KairosVL achieves a remarkable 61.8\% performance improvement. Compared to existing MLLMs, KairosVL maintains competitive performance among top-tier models despite having only 7B parameters. Furthermore, the model trained with two rounds of foundational capability-enhancing RL consistently outperforms the model trained with only one round of RL, demonstrating the necessity of the first round perception enhancement. Also, on this zero-shot dataset, KairosVL trained with RL consistently outperform the one trained with SFT, indicating that the two-round RL training strategy endows the model with stronger generalization capability.

On Dataset $\mathcal{B}$, KairosVL consistently achieves performance improvements of over 30\% compared to the Base Model, achieving state-of-the-art results in Fact-Adherent tasks, Predictive tasks, and Event-Aware tasks. However, for Counterfactual tasks, KairosVL still lags behind Qwen2.5VL-72B and GPT-4o. We hypothesize that the limited model size constrains the base model's ability to form hypotheses about time series data and conduct further analysis, thereby reducing the likelihood of sampling correct reasoning chains and ultimately limiting the reinforcement learning effect. 

Overall, the results demonstrate that our approach effectively enhances model performance across various semantic-conditional time series reasoning tasks, and successfully generalizes reasoning capabilities to real-world time series analysis scenarios.

\subsection{Ablation Study}
\label{sec:exphase}
To evaluate the impact of the training process and configurations on model performance, we conducted a series of ablation studies.

\subsubsection{Ablation study on two phase training.} 

\begin{figure}[htb]
    \centering
    \includegraphics[width=0.9\linewidth]{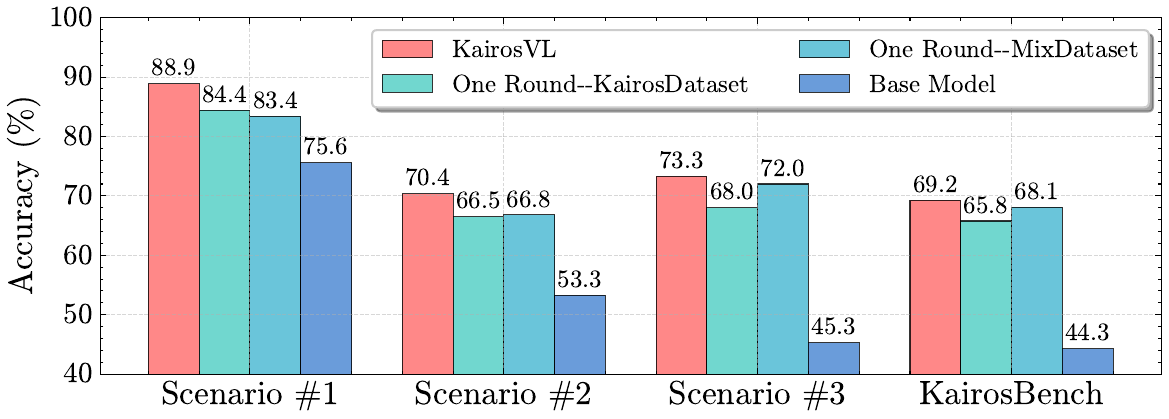}
    \caption{Ablation studies on different training pipeline.}
    \label{fig:phase}
\end{figure}

To verify the effectiveness of the two-round reinforcement learning scheme, we conducted an ablation study with two additional variants: (1) One-Round with KairosDataset, which performs a single round of reinforcement learning using only the KairosDataset; and (2) One-Round with MixDataset, which performs a single round of training using both the Primitive Dataset and KairosDataset. The latter is designed to test whether splitting the training process into multiple rounds can improve performance under the same data volume. As shown in Fig.~\ref{fig:phase}, our two-round training strategy consistently achieves the best results across different scenarios, demonstrating the importance of individual perception enhancement phase.

\begin{figure}[htb]
    \centering
    \subfloat[Response length during training.\label{fig:response}]{\includegraphics[width=.4\linewidth]{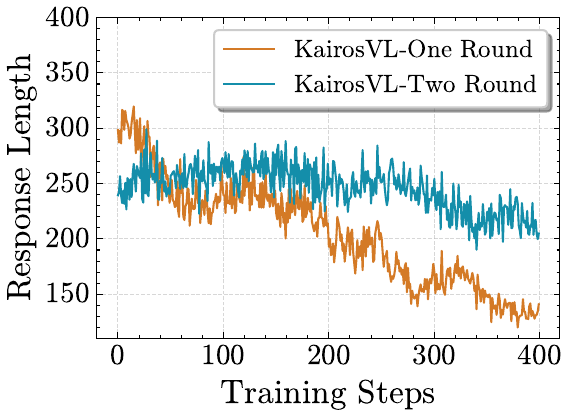}}
    \hspace{10pt}
	\subfloat[Entropy loss during training. \label{fig:entropy}]{\includegraphics[width=.4\linewidth]{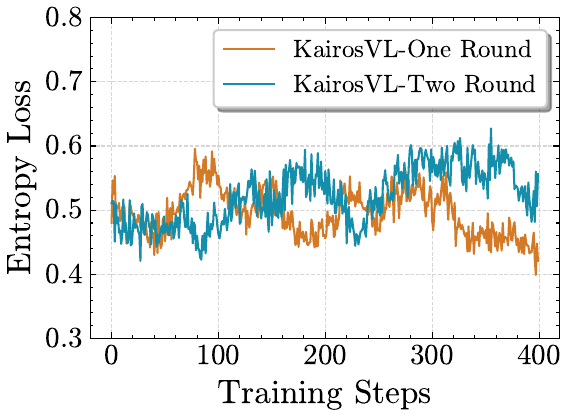}}
    \caption{Visualization of some metrics in training process.}
    \label{fig:training}
\end{figure}

We also visualize the changes in response length and entropy loss during both the one-round and two-round training processes (Fig.~\ref{fig:training}). As the number of training steps increases, the one-round variant exhibits a noticeable decline in both metrics, indicating that the model tends to produce less specific and more general answers. This behavior undermines the credibility of model decisions in practical applications.

\subsubsection{Ablation study on different KL settings.} 
\begin{figure}[htb]
    \centering
    \includegraphics[width=0.9\linewidth]{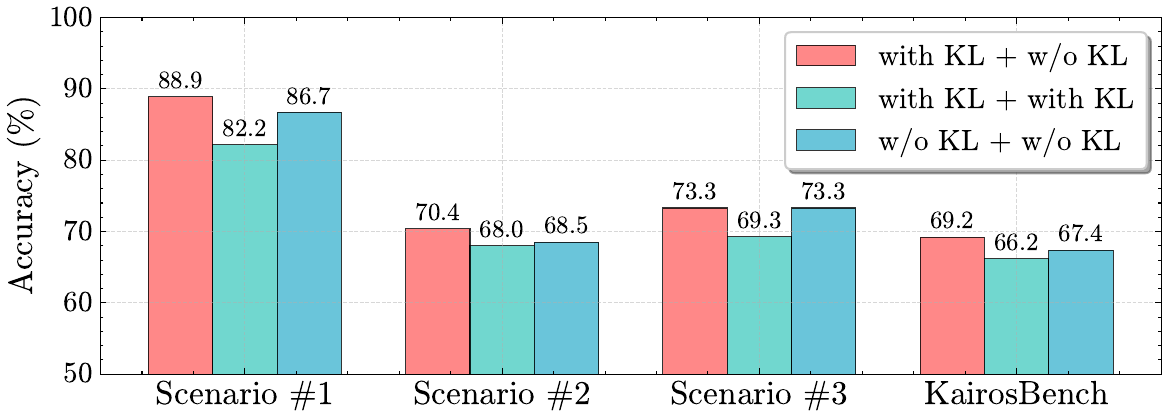}
    \caption{Ablation studies on different KL loss settings.}
    \label{fig:kl}
\end{figure}

Incorporating KL divergence in the training loss helps prevent the new policy from deviating excessively from the base model’s distribution, thus maintaining stability during optimization. However, it also constrains the model’s exploration ability during reinforcement learning. During the two-round training process, we also explored the impact of different KL-divergence settings on the experimental results. Specifically, as illustrated in Fig.~\ref{fig:kl}, we compared three configurations: introducing KL loss in both rounds, removing KL loss in both rounds, and introducing KL loss only in the first round while removing it in the second.  According to the experiment, the third setting achieves the consistent best performance. We conclude that introducing KL divergence during the perception enhancement stage is beneficial, as it stabilizes the learning of temporal concept alignment which doesn't rely on complex reasoning process. In contrast, removing KL divergence during the reasoning enhancement stage allows the model to explore more freely and develop diverse reasoning trajectories, leading to better downstream generalization and stronger overall reasoning performance.

\subsection{Case Study}
\label{sec:excase}
\begin{figure}[htb]
    \centering
    \includegraphics[width=\linewidth]{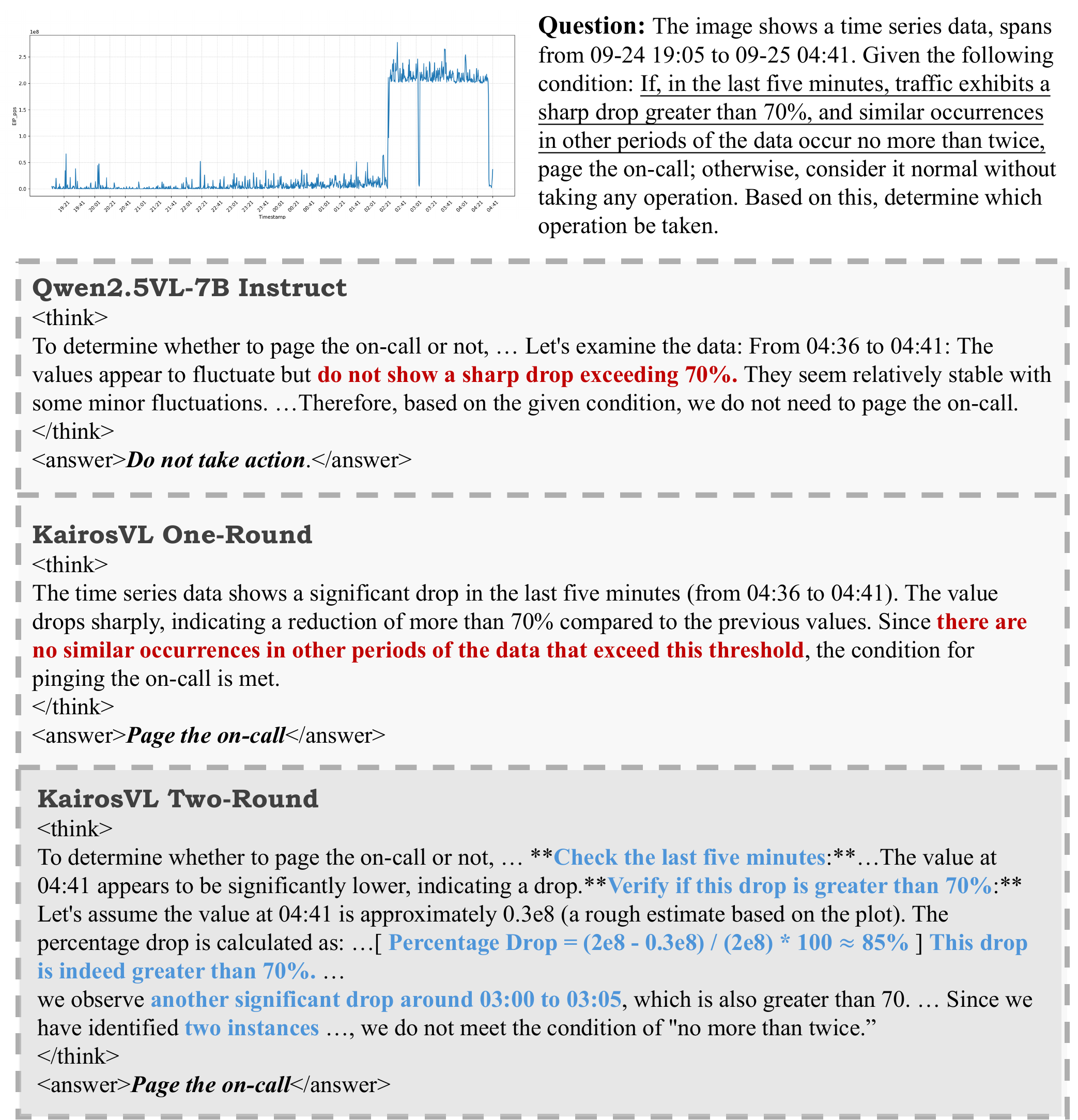}
    \caption{Case Study}
    \label{fig:case}
\end{figure}

As shown in the Fig.~\ref{fig:case}, we illustrate the performance of the Base Model and the reasoning-enhanced KairosVL on a real-world case. \textbf{The task requires the model to make a decision based on the SOP (the underlined text in the question) including two conditional factors: if the rate of decline $> 70\%$ and if the similar situations occur $\le2$ times within previous 8 hours}. Since traditional time series anomaly detection methods cannot understand what is "similar occurrences" described in the SOP, they can hardly work on this case.

From the case, we can observe that the Base Model makes an incorrect judgment regarding the rate of decline, leading to a wrong final decision. The KairosVL trained with only one round of RL produces a short and overly general response, and although it reached the correct decision, its reasoning process revealed a wrong judgment of the second condition. In contrast, the KairosVL trained with two rounds of RL not only generates a detailed and well-grounded reasoning process that integrates temporal patterns but also makes accurate judgments on both conditions. This demonstrates that KairosVL is capable of developing a structured and interpretable reasoning process, effectively aligning its decisions with real-world operational logic and improving the overall reliability of its temporal reasoning.

\section{Conclusion}
In this work, we address a critical gap in current time series modeling by introducing Semantic-Conditional Time Series Reasoning, a task that requires models to integrate temporal patterns with semantic and contextual information. We further propose a two-stage reinforcement learning framework that first strengthens the model’s perception of fundamental temporal primitives and then progressively enhances its reasoning capabilities. Our experiments demonstrate that this approach enables the model to produce more accurate and temporally grounded reasoning, while preserving intrinsic reasoning abilities. As an initial exploratory work, we hope it encourages the research community to pay greater attention to such semantic-conditional time series reasoning tasks, fostering the development of more advanced methods and broader applications for enhancing temporal reasoning which can satisfy the urgent demand in real-world applications.

\bibliographystyle{ACM-Reference-Format}
\bibliography{amain}

\appendix

\section{Comparison with Time Series Specific Models}
\label{app:tradits}
We additionally compare our method with representative time series–specific anomaly detection models~\cite{ano1,lstmad,mp,tranad} that take only raw time series as input on anomaly detection and mitigation oriented dataset $\mathcal{A}$. These models are designed to detect anomalies purely from numerical patterns and are unable to process natural-language SOPs or semantic descriptions that define when a behavior should be considered operationally abnormal.

In contrast to LLM/MLLM-based approaches which operates in a zero-shot manner by jointly reasoning over time series plots and SOPs at inference time, time series–specific models do not generalize across metrics without retraining. To acquire anomaly detection capability, they must be trained separately (or collecting statistics like MatrixProfile) on historical data for each individual metric, learning a metric-specific notion of normality before being applied to a given case. Concretely, for Dataset $\mathcal{A}$, each time series model is trained using the historical monitoring data of the target metric, and then outputs an anomaly score for the corresponding case. When the anomaly score exceeds a predefined threshold, the case is classified as anomalous and an on-call/alert priority escalation request is triggered. The following table (~\ref{tab:tsm}) reports the accuracy of the actions triggered by these methods.

\begin{table}[htb]
\centering
\caption{Action Accuracy (\%) of Time-Series-Only Anomaly Detection Methods on Dataset $\mathcal{A}$.}
\label{tab:tsm}
\resizebox{\linewidth}{!}{
\begin{tabular}{c|ccc}
\toprule[1.2pt]
Method        & Scenario \#1    & Scenario \#2    & Scenario \#3    \\ \midrule[1.2pt]
Donut~\cite{ano1}        & 26.5 & 27.3 & 21.6 \\
LSTMAD~\cite{lstmad}       & 32.9 & 21.3 & 27.9 \\
MatrixProfile~\cite{mp} & 30.4 & 35.6 & 25.4 \\
TranAD~\cite{tranad}        & 34.8 & 23.2 & 25.6 \\ \midrule
KairosVL      & 88.9 & 70.4 & 73.3 \\ \bottomrule[1.2pt]
\end{tabular}%
}
\end{table}

As expected, the performance of time series–specific models on this task are significantly inferior to KairosVL and can be considered practically unusable. Since these models rely exclusively on numerical patterns, they primarily detect out-of-distribution behaviors in terms of magnitude, trend, or variance shifts, and then map the high anomaly score to on-call/alert priority escalation action. However, \textbf{many of these cases are described as normal behaviors according to the descriptions in SOP (e.g., drop of network latency is a good thing for users and should not trigger alert), so almost all traditional time series–specific models fail to give correct decisions on these cases}. 

Lacking access to natural-language rules and contextual descriptions, traditional time series models are fundamentally incapable of incorporating such semantic constraints into their decisions. As a result, these models tend to over-trigger alerts for numerically rare but semantically benign patterns, while failing to align their predictions with the actual on-call criteria. This limitation highlights the gap between detecting statistical irregularities and making operationally meaningful anomaly decisions, and this is also a representative application scenario that motivates our formulation of the Semantic-Conditional Time Series Reasoning task.

\section{Time Series as Text}
\label{app:text}
\begin{figure}[htb]
    \centering
    \includegraphics[width=\linewidth]{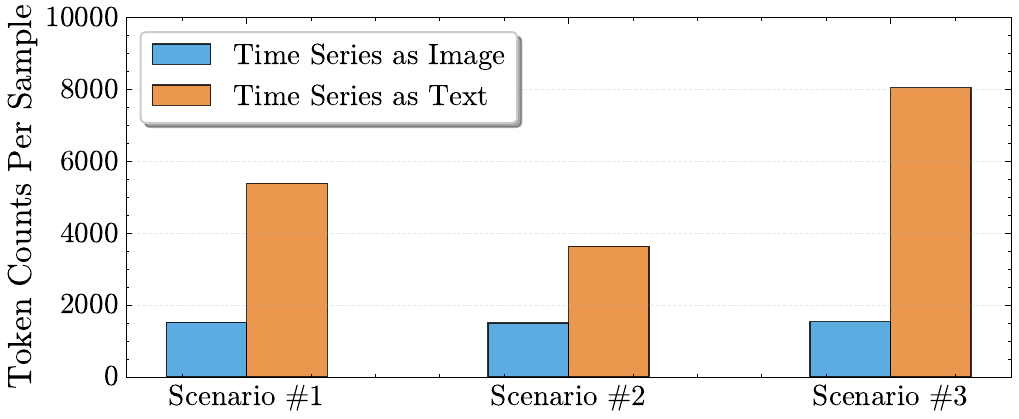}
    \caption{Comparison of token consumption between the “Time Series as Text” and “Time Series as Image” paradigms when using GPT-4o. Generally, when using GPT-4o, representing time series in textual form consumes about 5–8 tokens per numerical value. In contrast, even a high-resolution image of 5400 × 1800 pixels—which can visually encode tens of thousands of values—requires only around 1,400 tokens.}
    \label{fig:token}
\end{figure}

Apart from the existing consensus that representing time series as images generally outperforms representing them as text, we also evaluated the two approaches on commercial models in real-world scenarios. We found that when using time series as text, even with only a few hours of data sampled at 0.5-minute intervals, the token consumption becomes prohibitively high (Fig.~\ref{fig:token}). Moreover, during reasoning, the model often redundantly copies large amounts of data, resulting in excessively long responses, sometimes exceeding the model’s response length limit. Handling the data in text format also tends to produce more complex reasoning chains and data processing steps, which significantly reduces answer accuracy, as illustrated in Table~\ref{tab:textllm}. This can even cause the model to fail to produce valid answers (as observed with GLM-4V-Plus in Scenario \#3). In contrast, when representing time series as images—even at high resolution (5400 × 1800)—the model input only requires around 1,500 tokens, and the image can often encompass a longer time span of data. Therefore, representing time series as images provides a substantial cost advantage. Also, according to  the table, we observe that the Time Series as Image approach consistently and significantly outperforms the Time Series as Text approach across all scenarios. Taken together, these observations motivate our decision to enhance MLLMs’ temporal reasoning capabilities to better meet the demands of real-world time series analysis.

\begin{table}[htb]
\centering
\caption{Comparison of Question-Answer Accurary (\%) on real world time series reasoning tasks between "Time Series as Text" and "Time Series as Image" paradigms. Since each sample in Scenario \#3 contains a larger amount of data, GLM-4V-Plus struggles to produce valid responses within the limited output token budget when handling such long inputs. It often generates irrelevant gibberish or text unrelated to the question, resulting in an accuracy of zero for this scenario.}
\label{tab:textllm}
\begin{tabular}{c|ccc}
\toprule[1.5pt]
Models                   & Scen. \#1 & Scen. \#2 & Scen. \#3 \\ \midrule[1.5pt]
GPT-4o (text)            & 33.3         & 41.6         & 17.8         \\
GPT-4o (image)           & 83.3         & 64.5         & 77.3         \\ \midrule
Gemini-2.5 Flash (text)  & 72.6         & 42.5         & 17.3         \\
Gemini-2.5 Flash (image) & 79.4         & 62.5         & 48.7         \\ \midrule
GLM-4V-Plus (text)       & 38.9         & 30.8         & 0.0          \\
GLM-4V-Plus (image)      & 66.7         & 70.0         & 49.3         \\ \bottomrule[1.5pt]
\end{tabular}
\end{table}

\section{Existing Time Series Reasoning Tasks}
\label{app:task}
Recent research has begun to explore the capabilities of LLMs/MLLMs for time series reasoning, leading to the development of several new tasks and benchmarks. One branch~\cite{bench1-1, bench1-2, bench1-3, ben1-4, ben1-5} focused on testing models' ability to understand basic time series features, such as describing the overall trend (e.g., "upward"), detecting spikes, or recognizing seasonality. They effectively evaluate a model's basic comprehension but \textbf{do not probe their ability to perform the deeper, multi-step reasoning that builds upon these foundational features}. 

Another stream of work~\cite{bench2-1, bench2-2,bench2-4, tsaia, timera, timerbed, MCQ2, timemqa} focused on adapting traditional time series analysis samples to the format of answer generation under specific prompts and questions. For example, TimerBed~\cite{timerbed} convert time series classification tasks to multi-choice selections, while TSAIA~\cite{tsaia} includes some other tasks like prediction and anomaly detection. However, \textbf{the generated prompts often either present a time series in isolation, or additionally provide very limited context, making it impossible even for a human expert to arrive at a single unambiguous, verifiable answer}. Thus, the task is reduced to a form of pattern matching rather than genuine analysis, and the resulting "interpretations" are often superficial. Consequently, a significant gap remains between the tasks defined in existing benchmarks and the complex, contextual reasoning required in real-world analytical scenarios.

\begin{figure}[htb]
    \centering
    \includegraphics[width=\linewidth]{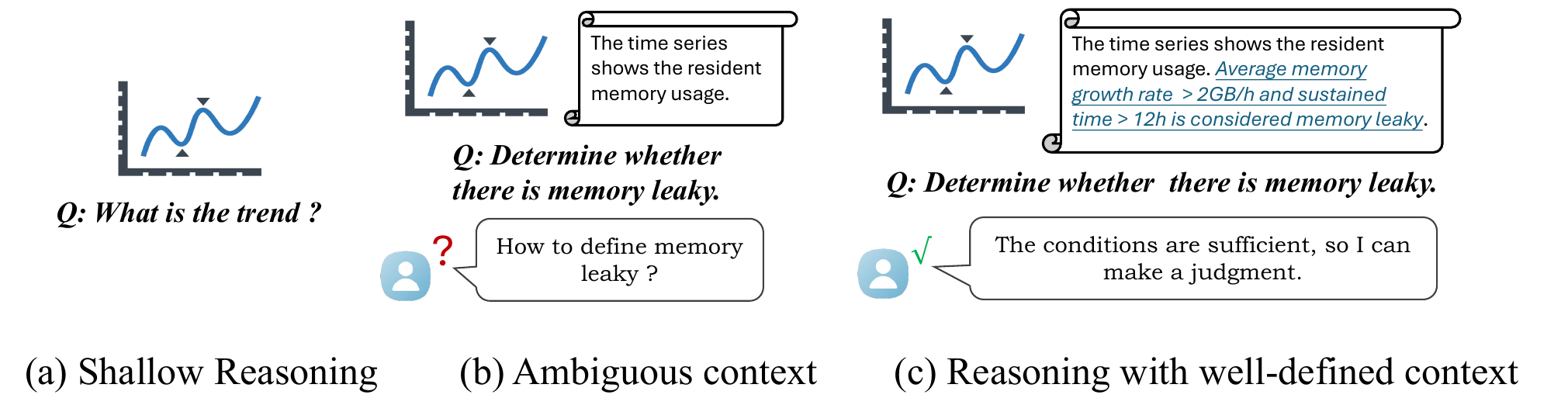}
    \caption{Drawbacks of existing time series reasoning tasks, and the ideal reasoning task.}
    \label{fig:taskcomp}
\end{figure}

\section{Details of Training with RLVR}
\label{app:train}
\subsection{Training Settings}
KairosVL is trained based on Qwen2.5VL-7B-Instruct~\cite{qwen2.5vl}. Both the Primitive Dataset and the KairosDataset consist of 2,000 samples, with the number of instances for each sub-task approximately balanced within each dataset. Full Parameter Reinforcement Fine-Tuning is used for KairosVL via veRL framework~\cite{verl,verl-easyr1} on the machine equipped with 112 Intel(R) Xeon(R) Platinum CPUs and 8 NVIDIA A800-SXM4-80G GPUs. 

For both two training phase, we conduct the full-parameter RFT on 4 GPUs. The rollout batch size is set 64, and for each prompt we rollout 4 samples. We use AdamW as the optimizer with learning rate set to $2\times10^{-7}$. During the first round perception enhancement reinforcement learning with verifiable reward, the clip ratio $\epsilon$ is set to 0.2, and we enable the KL loss item with coefficient $\beta$ set to $10^{-3}$. During the second round reasoning enhancement, we disable the KL loss item and increase the upper clip ratio $\epsilon'$ to 0.28.

\subsection{Rewards During Training Process}
\begin{figure}[htb]
    \centering
    \includegraphics[width=\linewidth]{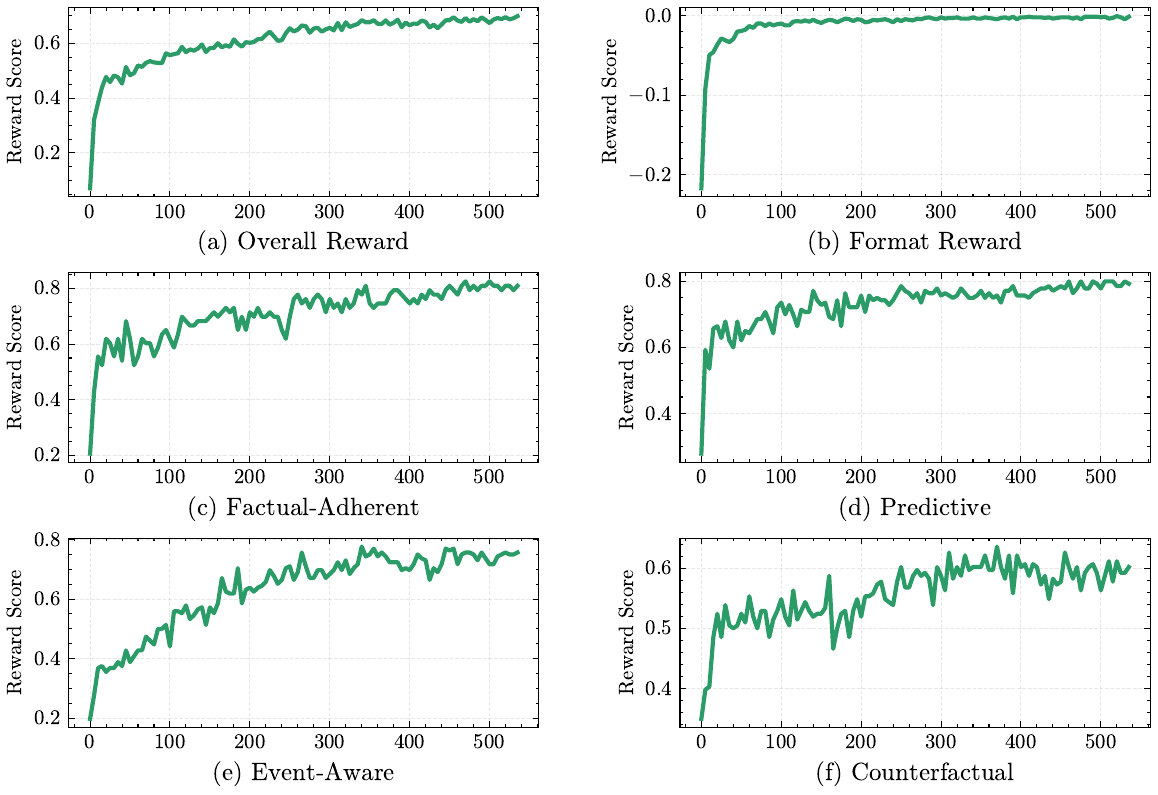}
    \caption{Rewards On Different Tasks During Training Process.}
    \label{fig:reward}
\end{figure}

Fig.~\ref{fig:reward} shows the rewards obtained on validation dataset during training across different tasks.

\section{The Experimental Margin of Error}
\label{app:margin}
Table~\ref{tab:mainres} shows the average accuracy of five samples for each question. Here, we report the KairosVL's margin of error of five experimental results relative to the mean.
\begin{table}[htb]
\centering
\caption{KairosVL's margin of error of five experimental results relative to the mean.}
\label{tab:margin}
\resizebox{0.6\linewidth}{!}{%
\begin{tabular}{c|c}
\toprule[1.2pt]
Dataset        & Margin of Error (\%) \\ \midrule
Scenario \#1   & $\pm0.9$             \\
Scenario \#2   & $\pm1.4$                  \\
Scenario \#3   & $\pm2.1$                  \\ \midrule
Fact-Adherent  & $\pm0.8$                  \\
Predictive     & $\pm0.6$                  \\
Event-Aware    & $\pm0.6$                  \\
Counterfactual & $\pm1.2$                  \\ \bottomrule[1.2pt]
\end{tabular}%
}
\end{table}

\section{Discussion}

\subsection{RLVR v.s. SFT}
\label{app:sft}
A key design choice in our framework is the adoption of Reinforcement Learning with Verifiable Reward (RLVR) instead of conventional Supervised Fine-Tuning (SFT). While SFT has been widely used to align large models with human-preferred behaviors, it presents several critical limitations when applied to complex time series reasoning tasks. When using direct question–answer pairs or pre-defined reasoning templates as SFT data, the model often becomes biased toward producing simplified or templated responses. This rigid supervision constrains the diversity of the model’s reasoning process, leading to loss of genuine chain-of-thought (CoT) behavior and weakened generalization to unseen scenarios. Moreover, constructing high-quality, diverse, and semantically consistent CoT datasets for time series reasoning is extremely labor-intensive and prone to human bias.

In contrast, RLVR enables the model to autonomously explore and optimize its reasoning process under verifiable reward constraints—requiring only outcome-level correctness rather than step-by-step human supervision. This paradigm not only significantly reduces the need for extensive human-crafted CoT data but also encourages the emergence of diverse reasoning trajectories that better capture the model’s intrinsic reasoning dynamics. This approach has been shown to achieve better performance and generalization ability compared with SFT in several works~\cite{sftrl1,sftrl2}, making it especially suitable for real-world time series reasoning tasks where interpretability, adaptability, and robustness are crucial.

\subsection{Potential Limitations}
Although KairosVL demonstrates strong performance across a wide range of time series reasoning tasks, several limitations remain that should be carefully considered when deploying the model in real-world applications.

\begin{itemize}
    \item \textbf{Inherent limitations of MLLMs.} Vision-based MLLMs may struggle with extremely fine-grained numerical distinctions (e.g., differentiating 150.1 from 150.01 on a scale of 100-200). Although they are already sufficient to meet the vast majority of real-world time series reasoning requirements, and they align well with the way humans perform time series analysis via visually experience and textual context, for the scenes requiring very accurate numerical computing and judgment, MLLM might not be the best choice.
    \item \textbf{Inadequate temporal localization.} We further observe in some failure cases that KairosVL sometimes struggles to precisely locate specific timestamps in a sequence. Instead of identifying the exact moment of interest, the model tends to approximate by referencing adjacent ticks or neighboring points in time. This behavior can lead to incorrect conclusions, especially in tasks where fine-grained temporal alignment is critical, such as diagnosing the immediate cause of a system outage or identifying the precise onset of an anomaly. This issue indicates that while the model grasps overall temporal structures, its understanding of fine temporal granularity remains imperfect.
    \item \textbf{Occasional reasoning hallucinations.} Another challenge lies in the model’s occasional generation of logically inconsistent reasoning paths. For instance, the model may produce erroneous comparisons such as asserting that 8,000 < 10,000 is false, or introduce unsupported causal links in the reasoning chain. Although such cases occur infrequently, they can lead to incorrect final decisions in safety- or reliability-critical scenarios. This reflects an inherent fragility in the model’s internal reasoning consistency and its ability to maintain factual alignment throughout multi-step inference.
\end{itemize}

These issues partly stem from the limitations imposed by the 7B model scale, which constrains the model’s capacity to perform extended chains of reasoning with high precision. Additionally, the current diversity and quantity of constructed time series reasoning tasks remain insufficient, which may introduce unintended inductive biases during reinforcement learning.

We call upon both the research and industrial communities to jointly advance this emerging area. Future efforts should focus on expanding the diversity and realism of time series reasoning datasets, developing scalable training paradigms that better integrate temporal and contextual understanding, and designing evaluation protocols that more accurately reflect the complexities of real-world decision-making. By addressing these challenges, we can collectively move toward more trustworthy, interpretable, and generalizable multimodal reasoning systems for time series intelligence.

\section{Prompt Instruction For Evaluation}
\label{app:prompt}
\begin{figure}[!h]
    \centering
    \includegraphics[width=\linewidth]{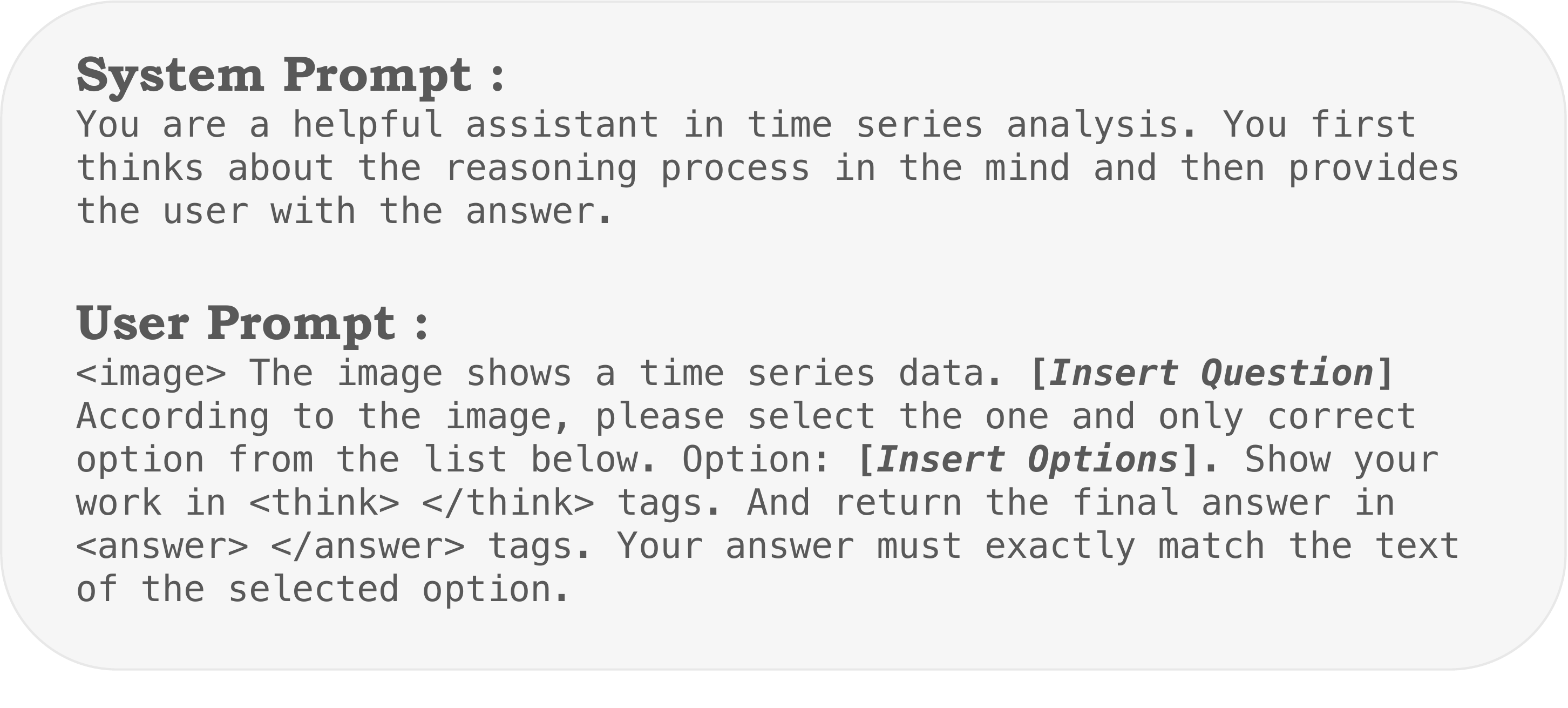}
    \caption{System and user prompts sent to MLLMs.}
    \label{fig:prompt}
\end{figure}

We prompt all models to exhibit thinking process before giving the final answers, as shown in Fig.~\ref{fig:prompt}.

\section{Prompt Instruction For KairosDataPipe}
\label{app:promptpipe}
\begin{figure*}[htb]
    \centering
    \subfloat[Fact-Adherent Task Description.\label{fig:dg1}]{\includegraphics[width=.95\linewidth]{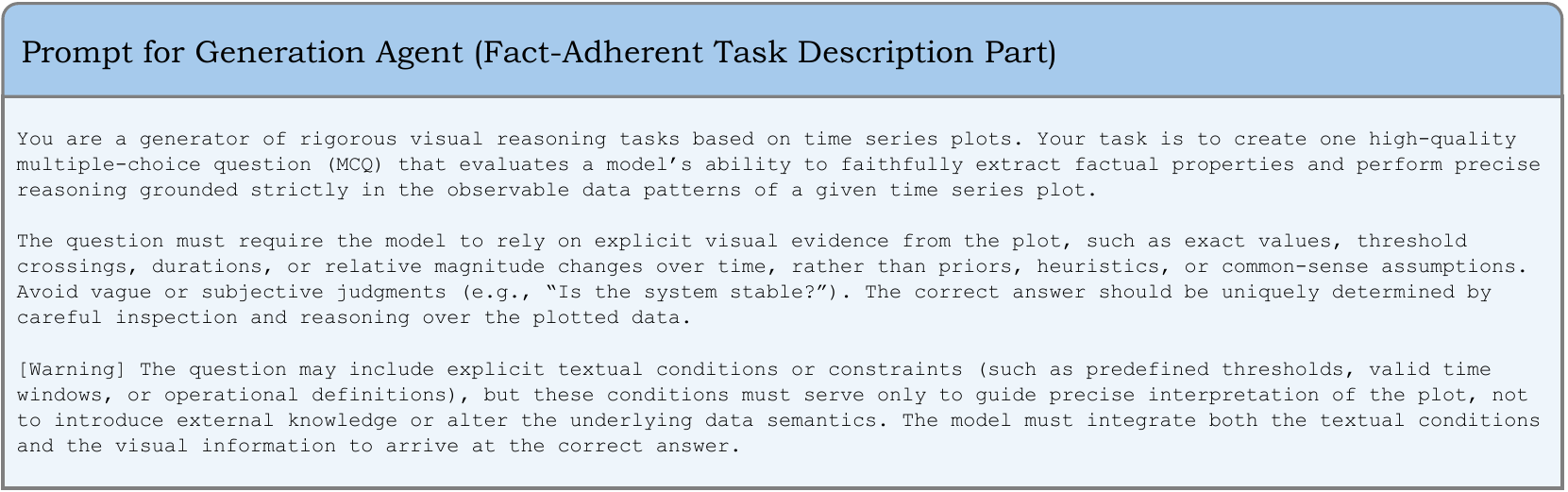}}

	\subfloat[Predictive Task Description.\label{fig:dg2}]{\includegraphics[width=.95\linewidth]{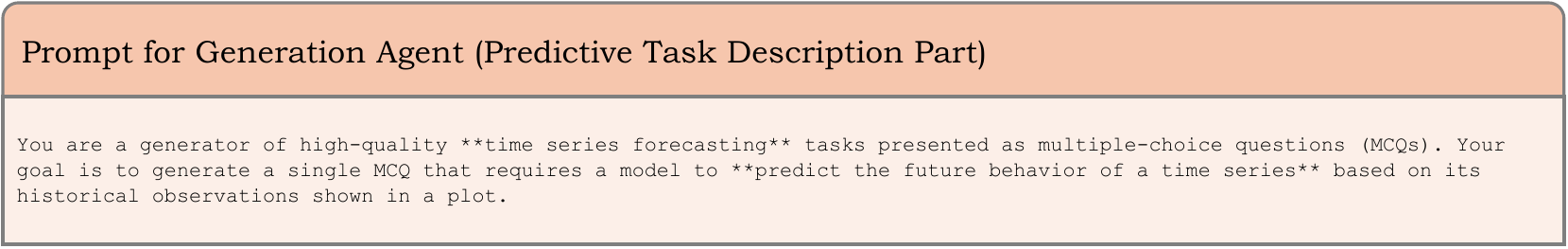}}

    \subfloat[Event-Aware Task Description.\label{fig:dg3}]{\includegraphics[width=.95\linewidth]{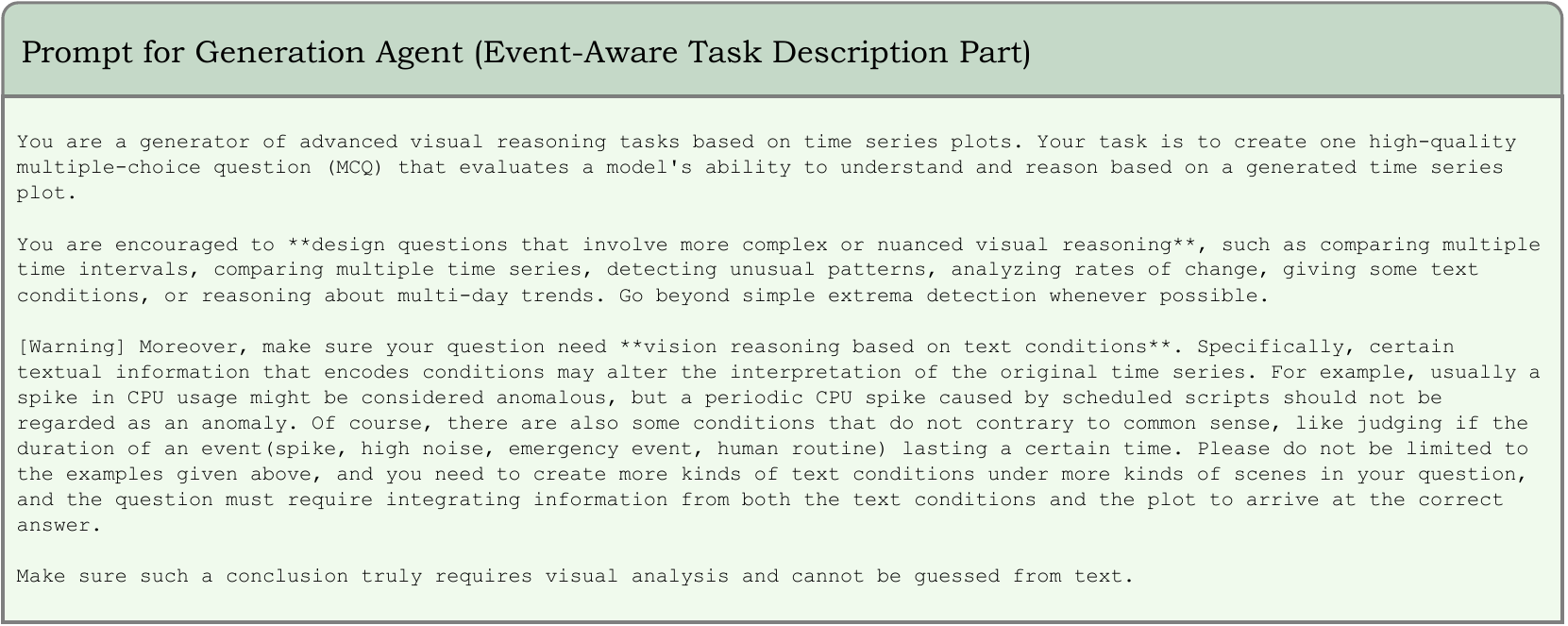}}
  
    \caption{The task specific descriptions for the Generation Agent (Part 1).}
    \label{fig:prompt_gen2}
\end{figure*}

\begin{figure*}[htb]
    \centering

    \subfloat[Counterfactual Task Description.\label{fig:dg4}]{\includegraphics[width=.95\linewidth]{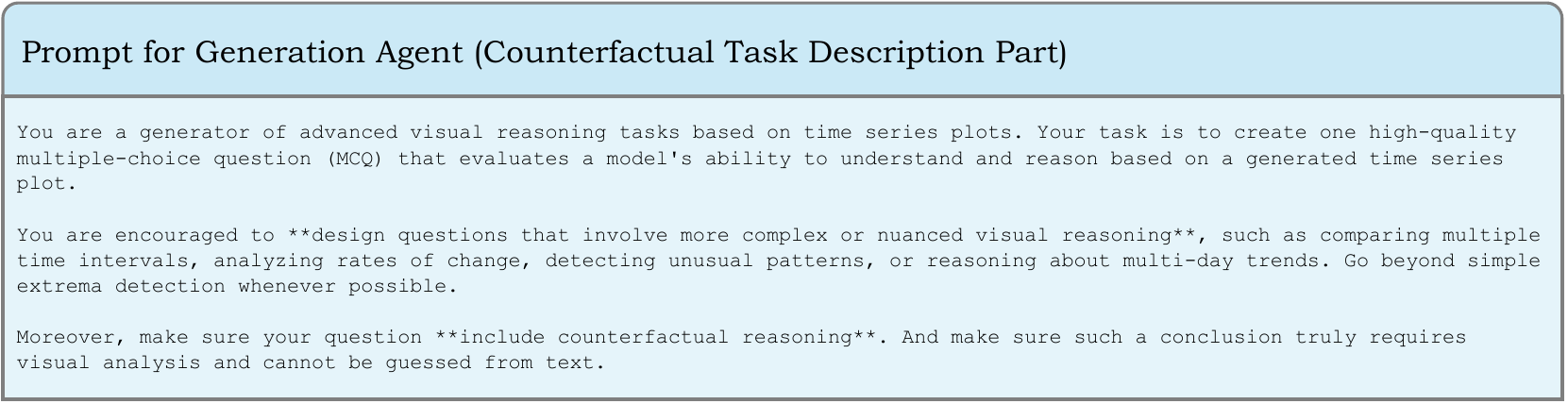}}
  
    \caption{The task specific descriptions for the Generation Agent (Part 2).}
    \label{fig:prompt_gen3}
\end{figure*}

\begin{figure*}[htb]
    \centering
    \includegraphics[width=0.95\linewidth]{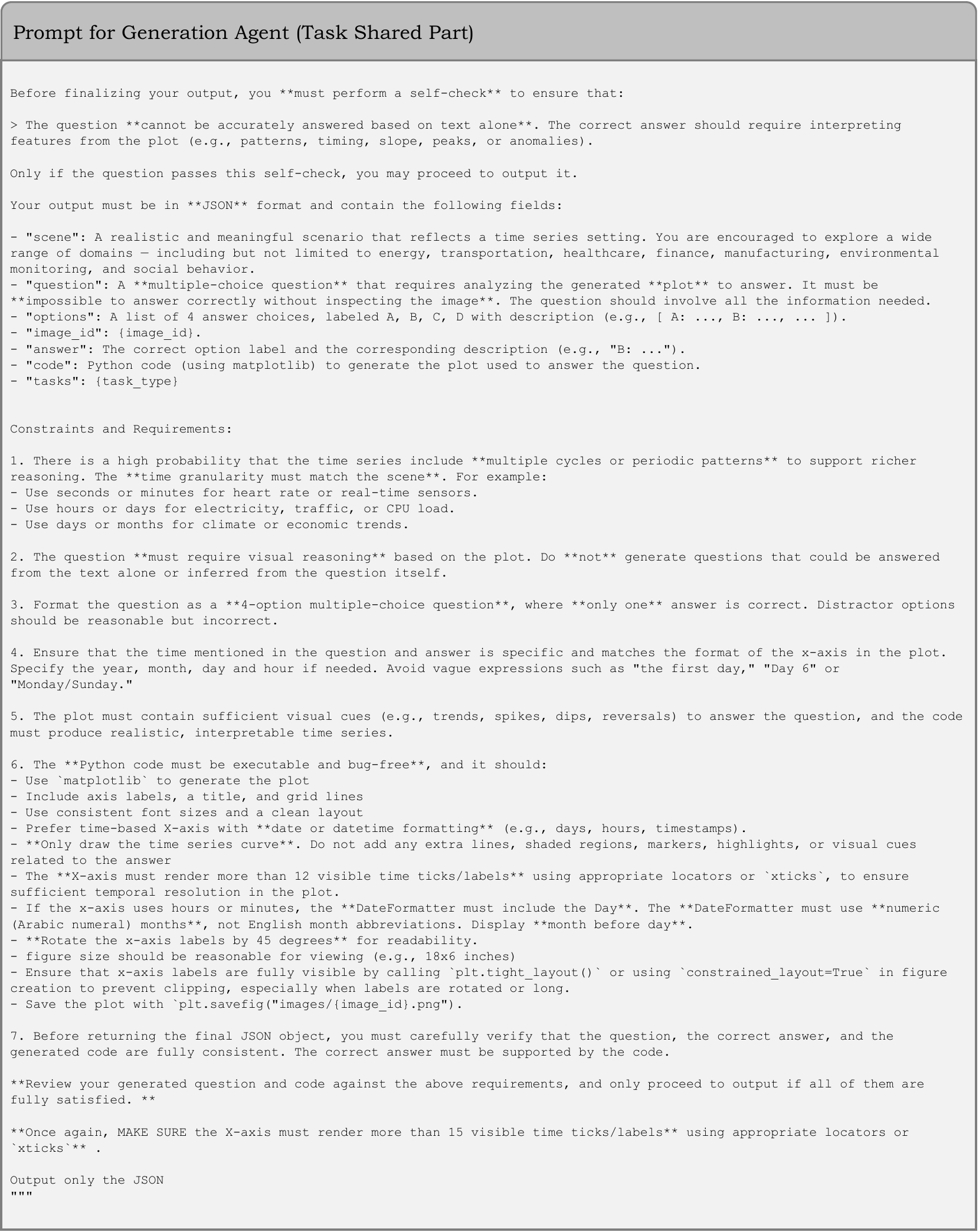}
    \caption{The task shared requirements for the Generation Agent.}
    \label{fig:prompt_gen1}
\end{figure*}

\begin{figure*}[htb]
    \centering
    \includegraphics[width=0.95\linewidth]{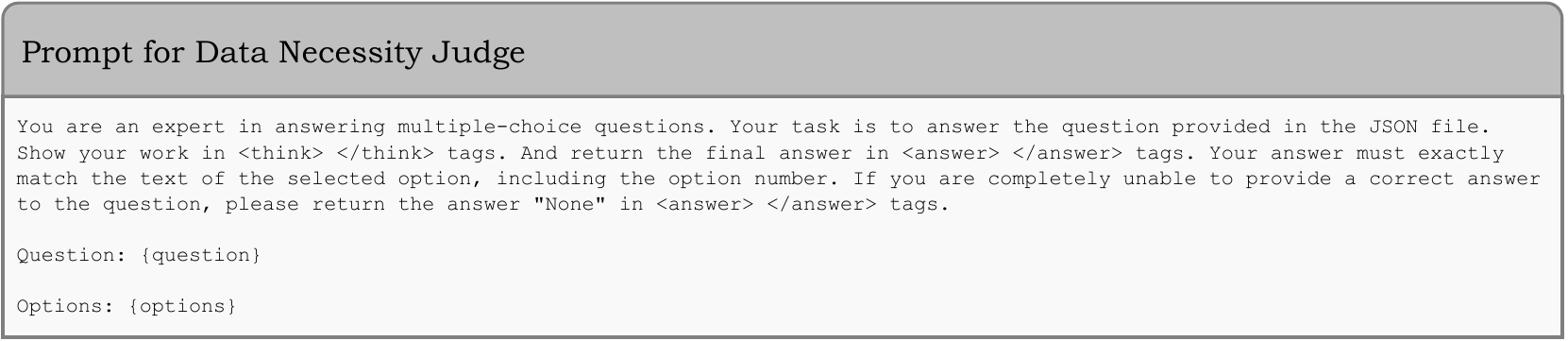}
    \caption{The prompt for the Data Necessity Judge.}
    \label{fig:ag2}
\end{figure*}

\begin{figure*}[htb]
    \centering
    \includegraphics[width=0.95\linewidth]{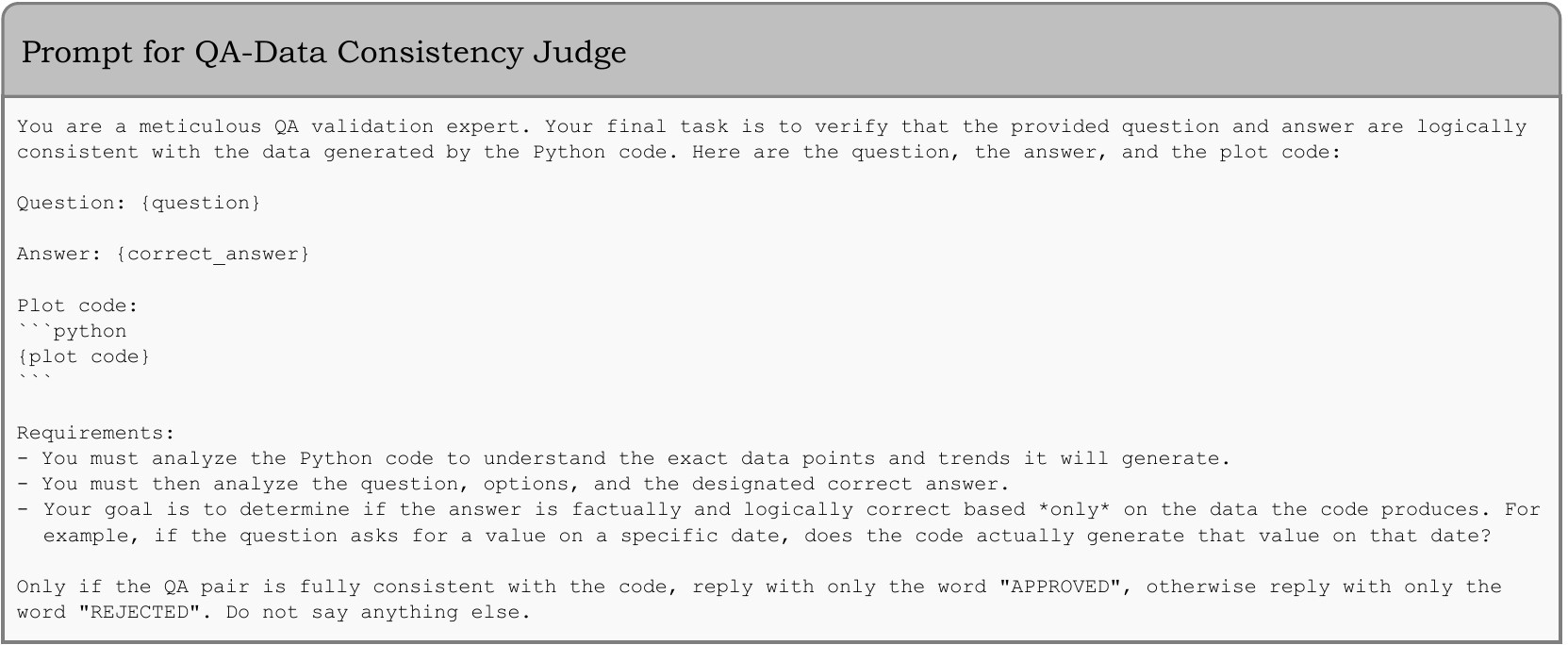}
    \caption{The prompt for the QA-Data Consistency Judge.}
    \label{fig:ag3}
\end{figure*}

\begin{figure*}[htb]
    \centering
    \includegraphics[width=0.95\linewidth]{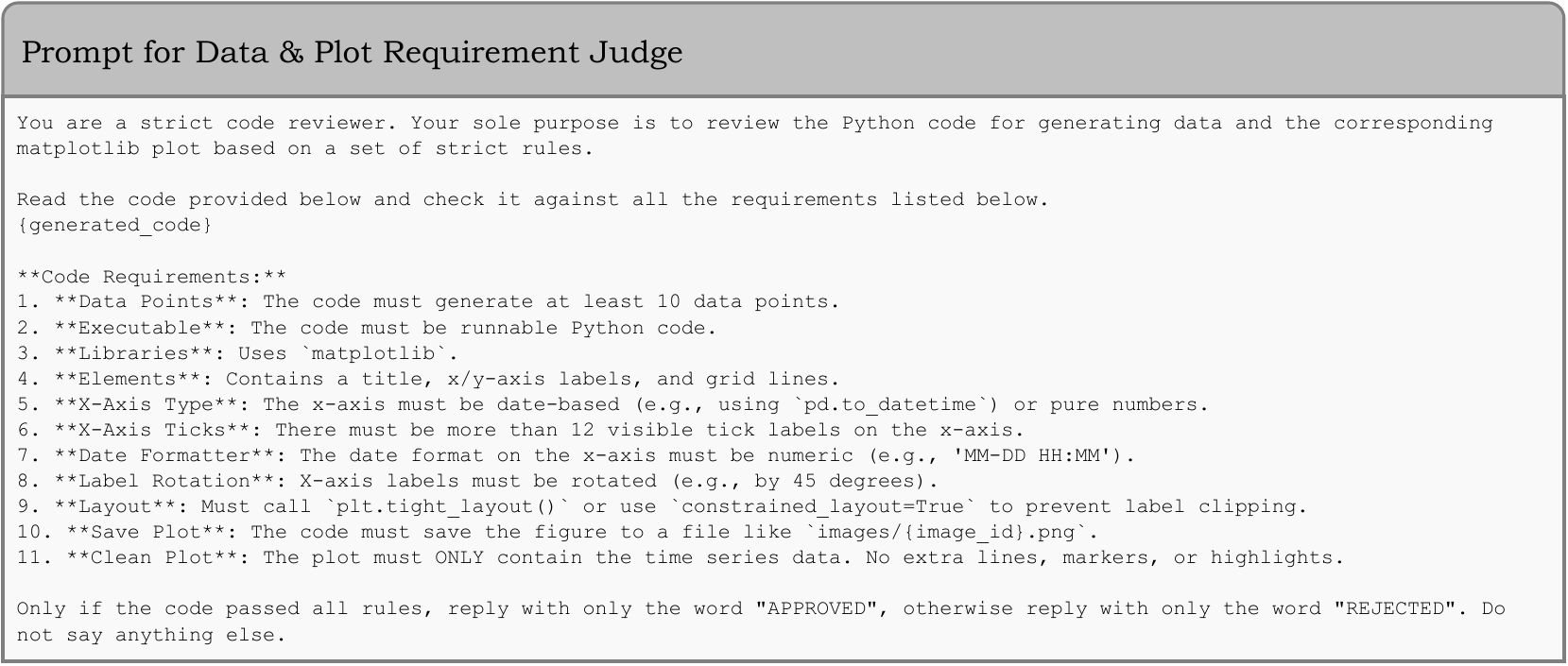}
    \caption{The prompt for the Data \& Plot Requirement Judge.}
    \label{fig:ag4}
\end{figure*}

In this section we provide prompts of all agents used to generate the training samples in KairosDataPipe. For the Generation Agent, we concatenate the task-specific description (Fig.~\ref{fig:prompt_gen2}, Fig.~\ref{fig:prompt_gen3}) and the shared requirements (Fig.~\ref{fig:prompt_gen1}) to generate prompts for each task, and the prompts are further sent to generation agent. For other judge agents (Fig.~\ref{fig:ag2}, Fig.~\ref{fig:ag3}, Fig.~\ref{fig:ag4}), the prompt is used to check all samples.

\section{Python Function for Executing Visualization Code}
\label{app:code1}
\begin{figure*}[htb]
    \centering
    \includegraphics[width=0.95\linewidth]{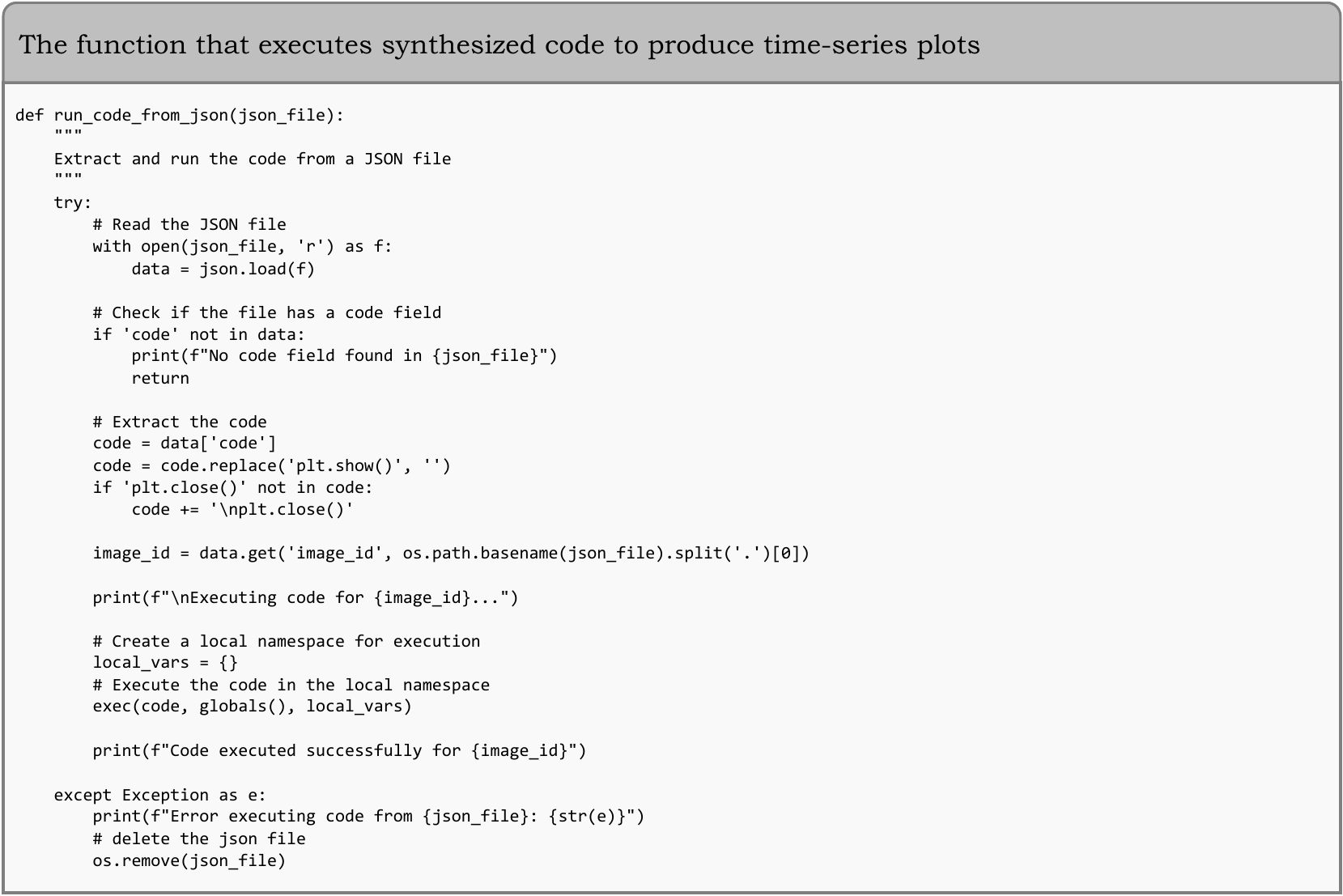}
    \caption{The python function that used to execute visualization code to generate time series plot.}
    \label{fig:code1}
\end{figure*}

The python function that used to execute visualization code to generate time series plot is shown in Fig.~\ref{fig:code1}.

\end{document}